\pdfoutput=1

\documentclass[11pt]{article}

\usepackage[preprint]{acl}

\usepackage{times}
\usepackage{latexsym}
\usepackage{comment}

\usepackage[T1]{fontenc}
\usepackage{fontawesome}
\usepackage[utf8]{inputenc}

\usepackage{microtype}
\usepackage{hyperref}
\usepackage{inconsolata}

\usepackage{graphicx}

\usepackage{listings}
\usepackage{xcolor,colortbl}
\usepackage[colorinlistoftodos,prependcaption,textsize=tiny]{todonotes}
\usepackage{amsmath}
\usepackage{amsfonts}
\usepackage{dsfont}
\usepackage{tcolorbox}

\usepackage[ruled,vlined,noend]{algorithm2e}
\SetKwInput{KwInput}{Input}
\SetKwInput{KwOutput}{Output}
\SetKwFor{For}{for}{:}{}
\SetKwFor{While}{while}{:}{}
\SetKwIF{If}{ElseIf}{Else}{if}{:}{elif}{else:}{}

\usepackage{mathtools} 

\usepackage{booktabs, multirow} 
\usepackage{soul}

\definecolor{codegreen}{rgb}{0,0.6,0}
\definecolor{codegray}{rgb}{0.5,0.5,0.5}
\definecolor{codepurple}{rgb}{0.58,0,0.82}
\definecolor{backcolour}{rgb}{0.95,0.95,0.92}

\definecolor{tartunlp_violet}{HTML}{7268D8}
\definecolor{tartunlp_violet_light}{HTML}{D3C6F7}

\definecolor{tartunlp_red}{HTML}{EF6650}
\definecolor{tartunlp_red_light}{HTML}{F7D4CB}

\definecolor{tartunlp_yellow}{HTML}{E0B12B}
\definecolor{tartunlp_yellow_light}{HTML}{FFDA99}

\definecolor{tartunlp_black}{HTML}{282828}
\definecolor{tartunlp_green}{HTML}{4DB6AC}
\definecolor{tartunlp_blue}{HTML}{3185FF}
\definecolor{tartunlp_gray}{HTML}{9B9B9B}

\lstdefinestyle{mystyle}{
    backgroundcolor=\color{backcolour},   
    commentstyle=\color{codegreen},
    keywordstyle=\color{magenta},
    numberstyle=\tiny\color{codegray},
    stringstyle=\color{codepurple},
    basicstyle=\ttfamily\footnotesize,
    breakatwhitespace=false,         
    breaklines=true,                 
    captionpos=b,                    
    keepspaces=true,                 
    numbers=left,                    
    numbersep=5pt,                  
    showspaces=false,                
    showstringspaces=false,
    showtabs=false,                  
    tabsize=2
}

\newcommand{\var}[1]{\textcolor{tartunlp_green}{\texttt{#1}}}
\newcommand{\kw}[1]{\textcolor{tartunlp_violet}{\textbf{#1}}}

\title{Teaching Old Tokenizers New Words:\protect\\ Efficient Tokenizer Adaptation for Pre-trained Models}

\author{Taido Purason\textsuperscript{1}, \, Pavel Chizhov\textsuperscript{2}, \, Ivan P. Yamshchikov\textsuperscript{2}, \, Mark Fishel\textsuperscript{1}
\\
\textsuperscript{1}Institute of Computer Science, University of Tartu
\\
\textsuperscript{2}CAIRO, Technical University of Applied Sciences Würzburg-Schweinfurt
\\
\texttt{taido.purason@ut.ee}
}

\begin{document}
\maketitle
\begin{abstract}
Tokenizer adaptation plays an important role in adapting pre-trained language models to new domains or languages. In this work, we address two complementary aspects of this process: vocabulary extension and pruning. The common approach to extension trains a new tokenizer on domain-specific text and appends the tokens that do not overlap with the existing vocabulary, which often results in many tokens that are unreachable or never used. We propose \textbf{continued BPE training} that extends a pre-trained tokenizer by continuing the BPE merge learning process on new data. Experiments across multiple languages and model families show that this approach improves tokenization efficiency and leads to better utilization of added vocabulary. We also introduce \textbf{leaf-based vocabulary pruning}, which removes redundant tokens while preserving model quality. Together, these methods provide practical tools for controlled vocabulary modification, which we release as an open-source toolkit.
\end{abstract}
\begin{center}
\faGithub~\footnotesize{\href{https://github.com/taidopurason/tokenizer-extension}{taidopurason/tokenizer-extension}}
\end{center}
\section{Introduction}
    When adapting large language models (LLMs) to new domains or languages, continued pre-training has become a widely used strategy. However, effective adaptation depends not only on updating model weights but also the tokenizer. Most LLMs rely on byte-pair encoding \cite[BPE,][]{sennrich-etal-2016-neural, gage1994new} tokenizers, and in practice their effectiveness varies substantially across languages. In particular, inefficient tokenization for underrepresented languages often leads to longer sequences and higher computational costs in LLMs \cite{petrov2023languagemodeltokenizersunfairness}. One way to address this is to extend the vocabulary with domain-specific tokens, which improves compression and reduces sequence lengths, albeit at the cost of increasing the model size due to a larger vocabulary. Pruning infrequent or irrelevant tokens can address this trade-off by reducing vocabulary size. In this work, we focus on tokenizer modification of pre-trained models using those two steps: pruning and extension.

\begin{figure}[t!]
    \centering
    \includegraphics[width=\linewidth]{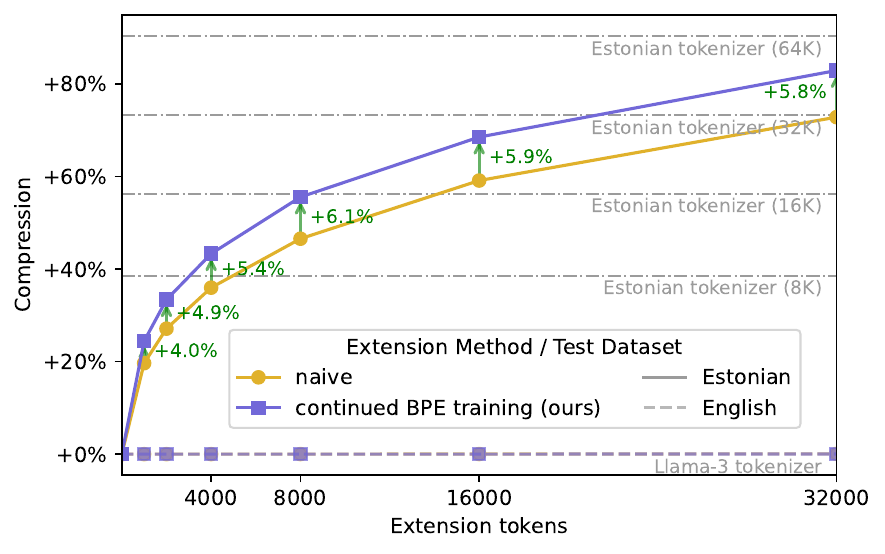}
    \caption{Change in text compression (\textit{bytes per token} $\uparrow$) of text using the Llama~3 tokenizer extended with Estonian tokens using naive tokenizer extension and continued BPE training (ours). We also compare to a tokenizer trained from scratch on Estonian data.}
    \label{fig:extension-compression}
\end{figure}

\begin{figure}[t!]
    \centering
    \includegraphics[width=0.95\linewidth]{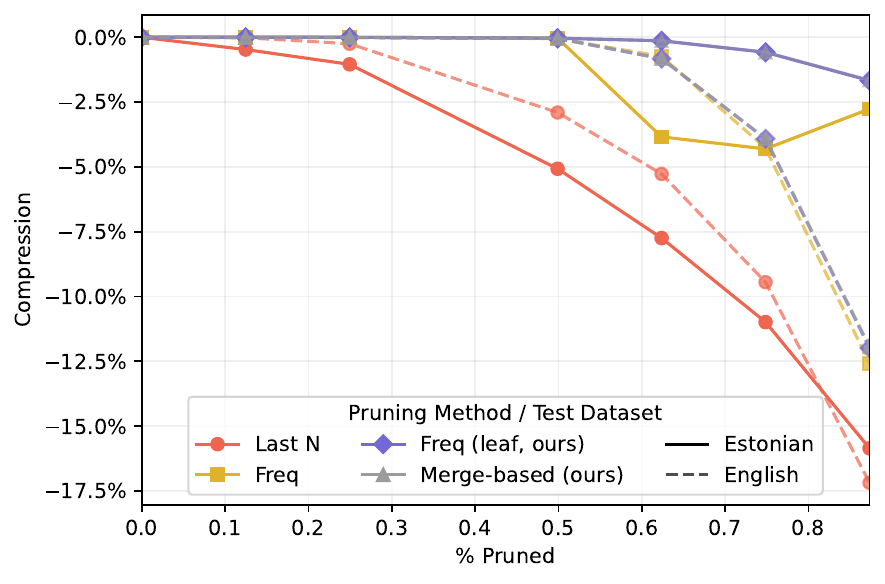}
    \caption{Change in text compression (\textit{bytes per token} $\uparrow$) for vocabulary pruning methods on Estonian-English pruning (Llama-3). Note that \textit{Freq (leaf)} and \textit{Merge-based} method results overlap.}
    \label{fig:prune-compression}
\end{figure}

\begin{figure}[!t]
    \centering
    \includegraphics[width=\linewidth]{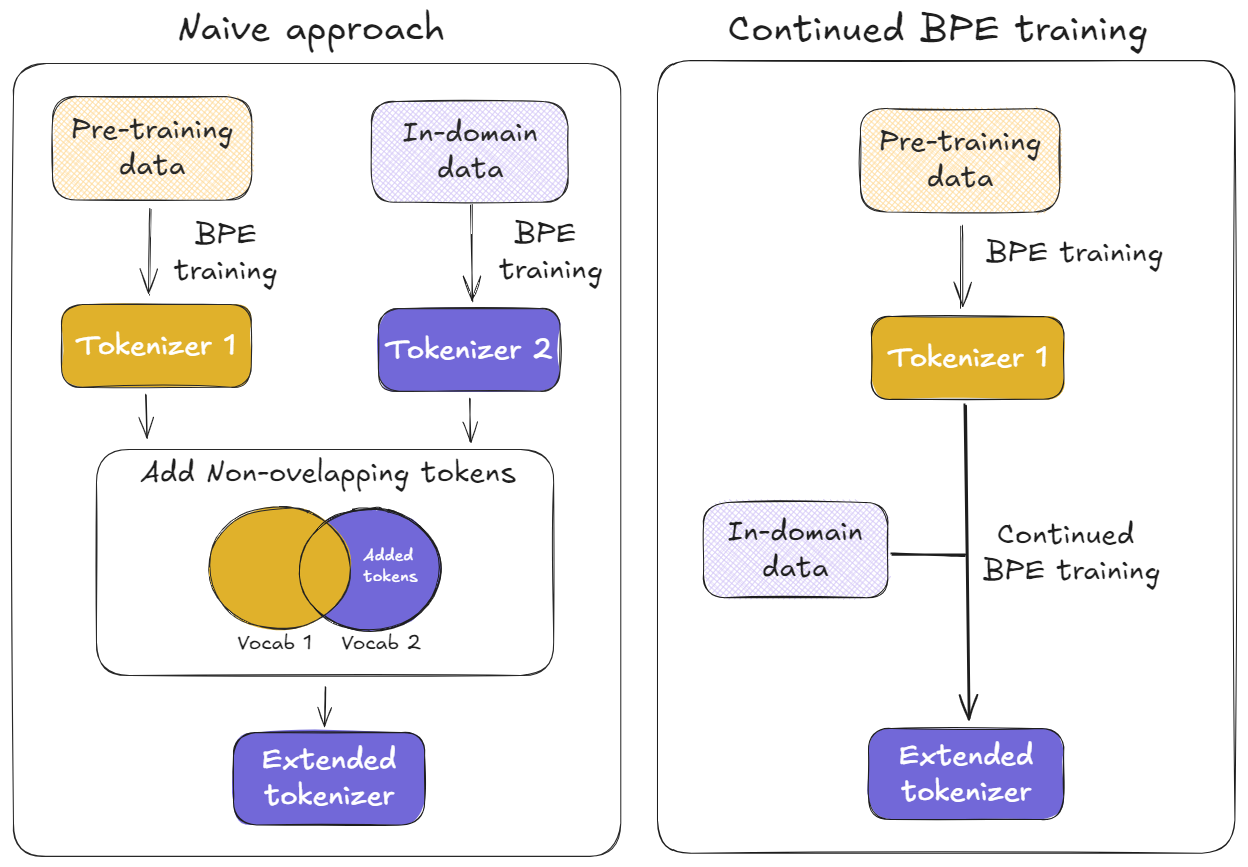}
    \caption{The comparison of vocabulary extension methods: previously common approach of training a tokenizer and using its vocabulary for extending an existing tokenizer (left) and continuing BPE training on an existing tokenizer proposed by us (right).}
    \label{fig:extension}
\end{figure}

 Extending an existing vocabulary is a widely used step in adapting pre-trained LLMs to new languages \cite{tejaswi-etal-2024-exploring, csaki-etal-2024-sambalingo, csaki2023efficientlyadaptingpretrainedlanguage, kiulian-etal-2025-english, fujii2024continual, lin2024mala500massivelanguageadaptation, yamaguchi2024effectivelyexpandvocabularyllms, cui2024efficienteffectivetextencoding}. In previous works, vocabulary extension typically involved training a new tokenizer on in-domain data and appending the tokens that were previously not present to the original tokenizer (see Figure~\ref{fig:extension}). We show that this approach introduces useless tokens that do not participate in merges, resulting in lower compression compared to having tokens that are useful in tokenization. This happens because BPE strictly follows the merge sequence, e.g., when a word \texttt{information} is originally tokenized as \texttt{inform + ation}, adding new tokens \texttt{infor} and \texttt{mation} is useless, even though they are new to the vocabulary. In this work, we propose a more effective alternative: continuing to train the original BPE tokenizer on domain-specific data. We show that for the same number of added tokens, this approach yields better compression (see Figure~\ref{fig:extension-compression}). While we focus on LLMs, our method can be applied to other pre-trained models that use BPE tokenizers.

The other side of tokenizer adaptation, besides extension, is pruning. Thus, we also study the impacts of different tokenizer pruning strategies and introduce a non-invasive leaf-based pruning method that trims the vocabulary in a structure-preserving manner.  This avoids creating unreachable tokens and yields higher tokenization efficiency (see Figure~\ref{fig:prune-compression}). By using token frequencies in a target corpus, akin to tokenizer training, we can preserve domain- and language-specific tokens without increasing sequence lengths in relevant domains while pruning irrelevant tokens. Our pruning method can also be used on its own in the context of model compression to reduce the vocabulary size, thereby substantially improving the model's efficiency: for instance, Gemma 3~\cite{gemmateam2025gemma3technicalreport} has 270 million parameters, 170 million of which are used for vocabulary embeddings~\cite{Lacombe2025Gemma3_270M}.

In summary, our contributions are threefold:
\begin{itemize}
    \item First, we introduce \textbf{a new method for vocabulary extension} based on continued BPE training, and provide a detailed analysis showing that it yields more efficient tokenization than the widely used approach of adding tokens from an independently trained domain-specific tokenizer. 
    \item Second, we develop \textbf{improved vocabulary pruning strategies} that do not change the tokenizer implementation and systematically analyze their effectiveness.
    \item Finally, \textbf{we release an open-source toolkit}\footnote{\href{https://github.com/taidopurason/tokenizer-extension}{https://github.com/taidopurason/tokenizer-extension}} that supports modification of both Byte-level and SentencePiece BPE tokenizers for pre-trained models through Hugging Face \texttt{transformers} \cite[HF,][]{wolf-etal-2020-transformers}. 
\end{itemize}

\section{Related work}
A growing body of work shows that tokenization quality is tied to model performance, highlighting its importance. Language-specific tokenizers have been shown to provide advantages over multilingual ones \citep{rust-etal-2021-good}, and more efficient compression is associated with better downstream results \citep{goldman-etal-2024-unpacking}. Recent evidence also suggests that differences in tokenization quality may contribute to the performance gap observed between agglutinative and fusional languages \citep{arnett-bergen-2025-language}. 

\subsection{Tokenizer extension}
Several works adopted the strategy of training a new tokenizer on domain- or language-specific data and adding the resulting non-overlapping tokens to the original vocabulary \cite{tejaswi-etal-2024-exploring, csaki-etal-2024-sambalingo, csaki2023efficientlyadaptingpretrainedlanguage, kiulian-etal-2025-english, fujii2024continual, lin2024mala500massivelanguageadaptation, yamaguchi2024effectivelyexpandvocabularyllms, cui2024efficienteffectivetextencoding}, which we refer to as the \textbf{naive approach}. MEDVOC-LLM \cite{balde-etal-2025-evaluation} provides another way to add new tokens, such as those proposed by a newly trained, domain-specific tokenizer. Specifically, each new token is tokenized with the original tokenizer and then recreated by merging the constituent tokens from left to right, adding intermediate merges and scaffold tokens. AdaptBPE \cite{balde-etal-2024-adaptive} integrates new tokens via longest string matching before tokenization, but this alters the tokenization function itself.

In contrast to the aforementioned methods, rather than specifying the desired tokens in advance and trying to integrate them, \textbf{continued BPE training} (ours) directly continues the BPE training process from the pre-trained tokenizer on in-domain data. This ensures that all added tokens and merges are fully compatible and optimal under the BPE objective, without introducing any unreachable tokens, thereby solving both the problem of identifying the tokens to add and their optimal integration.

\subsection{Tokenizer pruning}

In addition to extension, we introduce improved pruning techniques that surpass naive frequency-based methods \cite{yang-etal-2022-textpruner, csaki2023efficientlyadaptingpretrainedlanguage}, allowing for more extensive and targeted pruning without compromising compression or downstream performance. Prior pruning methods, such as BPE-Trimming \cite{cognetta-etal-2024-analysis}, target low-frequency tokens and split them during inference. We extend this work by introducing leaf-based pruning and trimming the tokenizer itself so that the inference remains identical to the basic BPE.
BPE-Knockout \cite{bauwens-delobelle-2024-bpe} removes tokens that have little value in terms of morphology and introduces tuple merges to compensate for their absence.
PickyBPE \cite{chizhov-etal-2024-bpe} and Scaffold-BPE \cite{lian2025scaffold} represent a different line of work of refining the vocabulary during tokenizer training by removing the intermediate, ``scaffolding'' tokens. 
While our pruning method also removes tokens, it does so to eliminate unused tokens in application to a narrower set of languages or domains. Our approach therefore modifies only the vocabulary and merge list while leaving the tokenization function unchanged, maintaining full compatibility with the original implementation and enabling further extensions.

\subsection{Replacing the tokenizer}
Instead of pruning and extending, previous works have also successfully replaced the previous tokenizer with a new one entirely \cite{samuel-etal-2025-small}. However, \citet{dagan2024getting} show that replacing or significantly altering the tokenizer requires extensive continued pre-training to avoid performance degradation, and \citet{tejaswi-etal-2024-exploring} find that the larger the vocabulary extension, the longer training is needed. We leave the question of what is the best strategy---modifying or replacing the tokenizer---to future research.

\subsection{Modification of embeddings}
When the tokenizer vocabulary is modified, the model embeddings must be updated accordingly. \citet{tejaswi-etal-2024-exploring} and \citet{csaki-etal-2024-sambalingo} find that Fast Vocabulary Transfer \cite[FVT,][]{gee-etal-2022-fast}, which copies embeddings for overlapping tokens and initializes new ones by averaging those of their decompositions under the original tokenizer (see Appendix~\ref{sec:fvt}), performs competitively among initialization strategies. While alternative methods exist \cite{dobler-de-melo-2023-focus, samuel-etal-2025-small}, our work focuses on selecting which tokens to add or prune at the tokenizer level rather than on embedding initialization.

\section{Methodology}
\subsection{Vocabulary extension}

\paragraph{Naive extension} It is common to extend a tokenizer with new tokens by training a new tokenizer with in-domain data and adding the non-overlapping tokens in the order of the trained vocabulary. As HF transformers relies on merge lists, it is also necessary to create merges leading to the added tokens. We generate the merges leading to the new tokens based on the token priorities, as opposed to simply merging the two tokenizers' merge lists, since we found the prior provides better results (see Appendix~\ref{sec:naive-extension-methods}). We refer to this method as \textbf{the naive approach}.

\paragraph{Continued tokenizer training} As an improvement, we propose \textbf{continued tokenizer training} to improve tokenization efficiency (see Figure~\ref{fig:extension}). We define \textbf{continued BPE training} as reapplying the BPE tokenizer training algorithm to in-domain data, using token pair frequencies of the text tokenized with the existing tokenizer to create the new merges. During merge creation, we ensure that each resulting token is valid according to the training implementation of the underlying tokenizer.

\textbf{For Byte-level BPE models} \cite[BBPE][]{radford2019language}, this involves first applying the pre-tokenizer to segment the text, then tokenizing the resulting segments to count token pair frequencies within the pre-tokens.

\textbf{For SentencePiece BPE models} \cite{kudo-richardson-2018-sentencepiece}, we tokenize the text and follow SentencePiece's merging rules by respecting constraints such as keeping different scripts separate and only allowing the space symbol at the beginning of a token.  If necessary, we also extend the vocabulary with missing characters to ensure coverage before finding extension merges and ensure that the training data is Unicode NFKC normalized.

In both cases, we then apply the standard BPE algorithm for tokenizer training using pair frequencies from the tokenized texts and creating merges until the desired number of new tokens is reached. For extending the original tokenizer we simply append the new tokens to the vocabulary and the new merges to the merge list in the order they were created. Similar to BPE tokenizer training, this method is language-agnostic and only requires texts in the target language.


\subsection{Vocabulary Pruning}

To allow for controlled reduction of vocabulary size, we also implement vocabulary pruning. This process involves removing tokens and the merges that lead to them, freeing space for vocabulary extension. To determine the tokens that should be pruned first, we try several approaches, based either on token ID (Last N) or on token frequency in a representative text sample. When implemented naively, both approaches might lead to unreachable tokens: IDs do not always reflect the order of merges, as in Llama 3.2~\cite{grattafiori2024llama3herdmodels}, and intermediate tokens tend to have lower frequencies than their downstream full-word tokens~\cite{chizhov-etal-2024-bpe}.

Addressing this issue, we implemented a \textbf{leaf-based pruning} algorithm, which iteratively collects the pruning order of tokens that are leaves in the BPE merge graph, i.e., they do not participate in merges and thus do not produce any downstream tokens. In Algorithm~\ref{alg:leaf_frequency_pruning}, we show pseudocode for frequency-based pruning of leaf tokens. First, atomic and leaf tokens are selected using the merge list. Leaf tokens are used to populate the priority queue, while atomic tokens cannot be removed and should never be added to the queue. After this, on each iteration, the lowest-priority element is taken from the queue and split into its merging sub-tokens. For each of the subtokens, we check the number of downstream merges it participates in, and, if there are no downstream merges left, this token is also considered a leaf and added to the priority queue for removal. The process ends when the queue runs out of leaves, which is the point when all non-atomic tokens are split.

We also introduce a merge-based pruning algorithm, which yields results similar to leaf-based frequency pruning. As its performance closely matches that of leaf-based pruning, we provide a detailed description in Appendix~\ref{sec:merge-based-pruning}.

\begin{algorithm}[!t]
\small 
\DontPrintSemicolon
\SetAlgoLined
\KwInput{%
\\
  \Indp\var{vocab}: mapping ``token $\to$ vocabulary index'' \\
  \var{unreachable}: set of unreachable tokens \\
  \var{merges}: list of BPE merges \\
  \var{counts}: mapping ``token $\to$ frequency in the target corpus''
}
\KwOutput{%
\\
\Indp
  \var{prune\_order}: list of tokens to prune in the order of pruning.
}
\;
\kw{initialize} \var{atomics} $\gets$ set of tokens not reachable by merges\;
\kw{initialize} \var{leaves} $\gets$ set of merged tokens not participating in merges + \var{unreachable}\;
\kw{initialize} \var{downstream\_merges} $\gets$ mapping ``token $\to$ number of merges with it''\;
\kw{initialize} \var{token\_splits} $\gets$ mapping ``token $\to$ merge leading to it''\;
\kw{initialize} \var{prune\_order} $\gets$ []\;
\kw{initialize} \var{queue} $\gets$ heap priority queue ``leaf $\gets$ (frequency, index)''\;

\vspace{0.3em}
\While{\var{queue}}{
    \var{token} $\gets$ \var{queue}.popmin()\;
    \var{pruning\_order}.append(\var{token})\;
    $(t_1, t_2) \gets$ \var{token\_splits}[\var{token}]\;
    \kw{update} $t_1, t_2$ in \var{queue} by \var{queue}[\var{token}]\;
    \For{$t \in \{t_1, t_2\}$}{
        \var{downstream\_merges}$[\,t\,] \mathrel{-}= 1$\;
        \If{$($\var{downstream\_merges}$[\,t\,] == 0)$ $\And (t \notin$ \var{atomics}$)$ }{
            add $t$ to \var{leaves} and \var{queue}\;
        }
    }
}
\vspace{0.3em}
\textbf{return} \var{pruning\_order}\;

\caption{Frequency Leaf Pruning}\label{alg:leaf_frequency_pruning}
\end{algorithm}

\subsection{Detecting unreachable tokens}
\label{sec:detecting-unreach-tokens}
We extend the method of \citet{land2024fishing} for identifying unreachable tokens to all vocabulary items (including those not decodable to UTF-8) and provide a formal definition. 
We define the \textbf{Self-Tokenization Test} (\textbf{STT}) as \textbf{the number of tokens unreachable through merges}:
\[
STT = \sum_{t \in \mathcal{V}} \mathds{1}\left[ tokenize(t) \neq [t] \right]
\]
The main idea is that if tokenizing a vocabulary token with the same tokenizer\footnote{In practice, this means that we test the raw string corresponding to the token, bypassing pre-processing, and disabling merge skipping (\texttt{ignore\_merges}).} does not reproduce that token, then the token cannot be formed through merge operations\footnote{It may still be produced through merge skipping.}.

\paragraph{Rationale.}
Let a BPE tokenizer be defined by a vocabulary $\mathcal{V}$ and a deterministic sequence of local merges, applied without merge skipping. For a token $t \in \mathcal{V}$, if $\text{tokenize}(t) \neq [t]$, then $t$ is \textbf{unreachable} by merges. If $\text{tokenize}(t) = [t]$, then $t$ could be reachable.

Suppose $t$ appears in the tokenized output of some pre-tokenized input $\mathcal{I}$. Then there must exist a substring of $\mathcal{I}$ equal to the literal form of $t$ that is merged into $t$. Because merges are local and merge skipping is disabled, the same sequence of merges must also apply when tokenizing the isolated string $t$. Therefore, if $\text{tokenize}(t) \neq [t]$, no input string can ever produce $t$, and $t$ is unreachable. However, if $\text{tokenize}(t) = [t]$, then $t$ is in principle reachable by merges assuming the pre-tokenizer produces $t$ as a substring to some input.

\paragraph{Merge Skipping.} If a pre-token output by a pre-tokenizer fully matches a vocabulary token, the merging process is skipped, and the token is produced immediately when merge skipping is enabled (\texttt{ignore\_merges} in HF). When applying this method, merge skipping is disabled because otherwise any existing vocabulary token trivially returns $[t]$, making the test meaningless. Some tokenizers (e.g., Llama~3) rely on this mechanism to produce certain tokens that are reachable only via merge skipping. While our method offers a fast check of merge-based reachability, its interpretation must account for whether tokens are produced via merge skipping in practice.

\section{Experimental setup}
\subsection{Datasets}
We use Fineweb-2 \cite[ODC-By license, multilingual]{penedo2025fineweb} and Fineweb \cite[ODC-By license, English]{NEURIPS2024_fineweb} as the training datasets and set a training budget of 100M characters for the comparison across 70 languages (see Table~\ref{tab:70-lang-overview} in Appendix~\ref{sec:language-overview}) and 1B characters for a more targeted evaluation for Estonian. We observe that increasing the amount of training data beyond this threshold does not yield significant improvements in tokenization efficiency during extension (see Appendix~\ref{sec:budget}).  

We frame the extrinsic evaluation as a case study on Estonian, using English-Estonian datamix for both tokenizer and LM training. Our setup reflects a common bilingual LLM adaptation scenario: extending a model to a new language while preserving its English capabilities. To this end, we allocated a total budget of 24B characters, equally divided between Estonian and English. The corpus was then tokenized using the different tokenizers under evaluation, ensuring that the underlying data remained constant across experiments involving changes to the tokenizer.

\subsection{Evaluation}
We evaluate compression on the FLORES-200 \cite{costa2022no} \texttt{devtest} split using \textit{bytes~per~token} ($BPT=\frac{\text{UTF-8 bytes}}{\text{tokens}}$) as the evaluation metric (higher is better). For comparison of different methods, we report the relative difference $\Delta_{rel}BPT = \frac{BPT_{ours}}{BPT_{naive}}-1$, which essentially states how many additional tokens the naive method produces compared to ours, as the number of bytes remains constant ($\frac{tokens_{naive}}{tokens_{ours}}-1$). We additionally report Rényi efficiency \cite[higher is better]{zouhar-etal-2023-tokenization}, which has been shown to correlate with downstream performance.

For downstream evaluation in Estonian and English, we evaluate on FLORES-200 \cite{nllb2022} (\textit{ET$\leftrightarrow$EN}), Winogrande \cite{sakaguchi2021winogrande}, Winogrande-ET \cite{ojastu2025estonianwinogrande}, XCOPA \cite{ponti-etal-2020-xcopa}, SIB200 \cite{adelani-etal-2024-sib}, and Belebele \cite{bandarkar-etal-2024-belebele} using \texttt{lm-evaluation-harness} \cite{eval-harness}. We evaluate FLORES-200 with COMET  \cite[\texttt{Unbabel/wmt22-comet-da},][]{rei-etal-2020-comet, rei-etal-2022-comet}. See Appendix~\ref{sec:evaluation-details} for more details.

\subsection{Models}

We look at four models Llama-3 \cite{grattafiori2024llama3herdmodels}, Llama-2 \cite{touvron2023llama2openfoundation}, Qwen-2.5 \cite{qwen2025qwen25technicalreport} and, Mistral Nemo\footnote{mistralai/Mistral-Nemo-Base-2407}. While Llama-3, Qwen-2.5, and Mistral Nemo use the BBPE tokenizer, while Llama-2 uses a SentencePiece BPE tokenizer. We use the tokenizers through Hugging Face \texttt{transformers} implementation \cite{wolf-etal-2020-transformers}. For experiments involving model training, we use Llama-3.2 1B and 3B and modify its embeddings after vocabulary modification with FVT \cite{gee-etal-2022-fast}. 

\section{Results}
\label{sec:results}
\subsection{Tokenizer evaluation}

\begin{table}[!t]\centering
\addtolength{\tabcolsep}{-0.7pt}
\scriptsize
\begin{tabular}{@{}l@{}cc@{\hspace{0.5\tabcolsep}}c@{}ccc@{}cccc@{}}\toprule
& &\multicolumn{2}{c}{\textbf{Compression $\uparrow$}} &\textbf{} &\multicolumn{2}{c}{\textbf{Rényi effic.}$\uparrow$} &\textbf{} &\multicolumn{2}{c}{\textbf{Unreach. added}$\downarrow$} \\\cmidrule{3-4}\cmidrule{6-7}\cmidrule{9-10}
\multicolumn{2}{c}{\textbf{Model}} &\textbf{$\Delta_{rel}$} &\textbf{WR} &\textbf{} &\textbf{$\Delta_{rel}$} &\textbf{WR} &\textbf{} &\textbf{ours} &\textbf{naive} \\\midrule
\multicolumn{10}{l}{\cellcolor[HTML]{f3f3f3}\textbf{Llama-3}} \\
&+1000 &\cellcolor[HTML]{f1ecfd}3.3\% &88.6\% & &\cellcolor[HTML]{fefbfb}-0.1\% &18.6\% & &0.0\% &\cellcolor[HTML]{fcede9}4.5\% \\
&+2000 &\cellcolor[HTML]{efeafc}3.6\% &92.9\% & &\cellcolor[HTML]{fdf4f2}-0.6\% &4.3\% & &0.0\% &\cellcolor[HTML]{fbe8e3}5.7\% \\
&+4000 &\cellcolor[HTML]{eee9fc}3.8\% &98.6\% & &\cellcolor[HTML]{fceeeb}-1.1\% &5.7\% & &0.0\% &\cellcolor[HTML]{fae3dd}7.0\% \\
&+8000 &\cellcolor[HTML]{eee9fc}3.7\% &100.0\% & &\cellcolor[HTML]{fbebe7}-1.4\% &1.4\% & &0.0\% &\cellcolor[HTML]{f9ded7}8.2\% \\
&+16000 &\cellcolor[HTML]{f0ecfd}3.3\% &98.6\% & &\cellcolor[HTML]{fbebe7}-1.3\% &1.4\% & &0.0\% &\cellcolor[HTML]{f8d9d1}9.3\% \\
&+32000 &\cellcolor[HTML]{f3f0fd}2.7\% &92.9\% & &\cellcolor[HTML]{fbeeeb}-1.1\% &7.1\% & &0.0\% &\cellcolor[HTML]{f8d6cd}10.2\% \\\midrule
\multicolumn{10}{l}{\cellcolor[HTML]{f3f3f3}\textbf{Qwen-2.5}} \\
&+1000 &\cellcolor[HTML]{ece6fc}4.3\% &98.6\% & &0.1\% &18.6\% & &0.0\% &\cellcolor[HTML]{fceeea}4.3\% \\
&+2000 &\cellcolor[HTML]{e6defb}5.6\% &97.1\% & &\cellcolor[HTML]{fcf0ed}-1.0\% &7.1\% & &0.0\% &\cellcolor[HTML]{fbe8e4}5.6\% \\
&+4000 &\cellcolor[HTML]{e0d6fa}6.9\% &100.0\% & &\cellcolor[HTML]{fae6e1}-1.8\% &2.9\% & &0.0\% &\cellcolor[HTML]{fae2dc}7.1\% \\
&+8000 &\cellcolor[HTML]{dacff9}8.1\% &98.6\% & &\cellcolor[HTML]{f8dcd5}-2.5\% &1.4\% & &0.0\% &\cellcolor[HTML]{f9ddd6}8.4\% \\
&+16000 &\cellcolor[HTML]{d6caf8}9.0\% &97.1\% & &\cellcolor[HTML]{f7d6ce}-3.0\% &1.4\% & &0.0\% &\cellcolor[HTML]{f8d8d0}9.6\% \\
&+32000 &\cellcolor[HTML]{d3c6f7}9.6\% &94.3\% & &\cellcolor[HTML]{f7d4cb}-3.2\% &8.6\% & &0.0\% &\cellcolor[HTML]{f7d4cb}10.5\% \\\midrule
\multicolumn{10}{l}{\cellcolor[HTML]{f3f3f3}\textbf{Mistral Nemo}} \\
&+1000 &\cellcolor[HTML]{f2edfd}3.0\% &97.1\% & &\cellcolor[HTML]{fdf6f4}-0.5\% &12.9\% & &0.0\% &\cellcolor[HTML]{fdf0ed}3.8\% \\
&+2000 &\cellcolor[HTML]{eee9fc}3.7\% &94.3\% & &\cellcolor[HTML]{fcf2f0}-0.8\% &5.7\% & &0.0\% &\cellcolor[HTML]{fcebe7}4.9\% \\
&+4000 &\cellcolor[HTML]{ece6fc}4.2\% &98.6\% & &\cellcolor[HTML]{fcefec}-1.0\% &4.3\% & &0.0\% &\cellcolor[HTML]{fbe6e0}6.3\% \\
&+8000 &\cellcolor[HTML]{ebe5fc}4.5\% &98.6\% & &\cellcolor[HTML]{fbeeea}-1.1\% &4.3\% & &0.0\% &\cellcolor[HTML]{fae1da}7.5\% \\
&+16000 &\cellcolor[HTML]{ebe4fc}4.6\% &100.0\% & &\cellcolor[HTML]{fcefeb}-1.1\% &4.3\% & &0.0\% &\cellcolor[HTML]{f9dcd4}8.7\% \\
&+32000 &\cellcolor[HTML]{ece6fc}4.3\% &98.6\% & &\cellcolor[HTML]{fcf1ef}-0.9\% &4.3\% & &0.0\% &\cellcolor[HTML]{f8d8cf}9.7\% \\\midrule
\multicolumn{10}{l}{\cellcolor[HTML]{f3f3f3}\textbf{Llama-2 (SentencePiece)}} \\
&+1000 &\cellcolor[HTML]{eae4fc}4.6\% &95.7\% & &\cellcolor[HTML]{fef9f8}-0.2\% &17.1\% & &0.0\% &\cellcolor[HTML]{fdf1ee}3.6\% \\
&+2000 &\cellcolor[HTML]{e8e1fb}5.2\% &95.7\% & &\cellcolor[HTML]{fcefeb}-1.1\% &8.6\% & &0.0\% &\cellcolor[HTML]{fcece8}4.8\% \\
&+4000 &\cellcolor[HTML]{e4dbfa}6.1\% &90.0\% & &\cellcolor[HTML]{fae6e1}-1.8\% &5.7\% & &0.0\% &\cellcolor[HTML]{fbe7e2}6.0\% \\
&+8000 &\cellcolor[HTML]{e0d6fa}6.9\% &84.3\% & &\cellcolor[HTML]{f9dfd8}-2.3\% &7.1\% & &0.0\% &\cellcolor[HTML]{fae3dd}7.0\% \\
&+16000 &\cellcolor[HTML]{ddd3f9}7.4\% &72.9\% & &\cellcolor[HTML]{f8dad2}-2.7\% &17.1\% & &0.0\% &\cellcolor[HTML]{fadfd9}7.8\% \\
&+32000 &\cellcolor[HTML]{dcd2f9}7.7\% &74.3\% & &\cellcolor[HTML]{f7d8d0}-2.8\% &21.4\% & &0.0\% &\cellcolor[HTML]{f9ddd6}8.4\% \\
\bottomrule
\end{tabular}
\caption[]{\textbf{Average scores across 70 languages}. We report the average relative gain ($\Delta_{rel}$) and Win Rate (\textit{WR}) of continued tokenizer training (ours) over extending from an independently trained tokenizer (naive) for compression (\textit{BPT}) and Rényi efficiency. We also report the percentage of added tokens that are unreachable through merges (\textit{STT}). \textit{WR} -- the percentage of languages where our method scores higher than the naive method.}\label{tab:70-lang-avg}
\end{table}

By continuing tokenizer training (ours) instead of extending a tokenizer from a new language-specific tokenizer (naive method), \textbf{we achieve up to 9.6\% higher tokenization efficiency on average}, with a vast majority (72.9\%--100\% depending on the tokenizer and extension size) of the 70 languages achieving higher compression (see Table~\ref{tab:70-lang-avg}). We do not observe any notable negative effects of our method on compression, while the increase can exceed 20\% compared to the naive method (see Appendix~\ref{sec:full-70-lang} for full results). Furthermore, there is also almost no effect on English tokenization after the target-language extension (see Table~\ref{tab:effect-on-eng} in Appendix~\ref{sec:effect-on-english}). Higher target-language compression means we can process the same text with fewer tokens during training and inference, consuming less computational resources.

\begin{figure}[!t]
    \centering
    \includegraphics[width=0.95\linewidth]{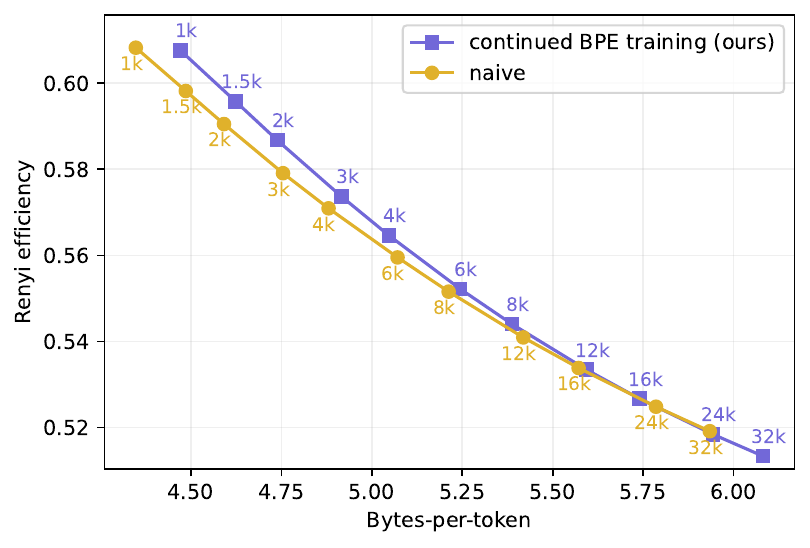}
    \caption{Llama-3 tokenizer Rényi efficiency $\uparrow$ vs compression (bytes per token) $\uparrow$ averaged over 70 languages. The point labels indicate the number of added tokens.}
    \label{fig:renyi-compression}
\end{figure}

When looking at Rényi efficiency (higher is better), there is a deterioration when comparing continued BPE training to naive extension. On a closer look, even though Rényi efficiency decreases for the model with the same number of added tokens, when looking at extended tokenizers with roughly equal compression, our method outperforms the naive method on average (see Figure~\ref{fig:renyi-compression}).

While our method is language-agnostic, we observe that the improvement over the naive method depends on the base tokenizer, language, and the number of added tokens. If we examine the average increase in compression of our method over the naive method, we see that it is more effective for languages with Cyrillic and Latin scripts (see Table~\ref{tab:70-lang-script}).
\begin{table}[!t]\centering
\addtolength{\tabcolsep}{-3pt}
\small
\begin{tabular}{@{}l@{\hspace{0.75\tabcolsep}}rccccc@{}}\toprule
& &\multicolumn{4}{c}{\textbf{$\Delta_{rel}$ Compression}$\uparrow$} \\\cmidrule{3-6}
\textbf{Script}&($n$)&\textbf{Llama-3} &\textbf{Qwen-2.5} &\textbf{Mistral} &\textbf{Llama-2} \\\midrule
Latn &(35) &\cellcolor[HTML]{ede8fc}4.7\% &\cellcolor[HTML]{d9cdf8}9.8\% &\cellcolor[HTML]{f3effd}3.3\% &\cellcolor[HTML]{d8ccf8}10.1\% \\
Cyrl &(9) &\cellcolor[HTML]{ece6fc}5.1\% &\cellcolor[HTML]{d3c6f7}11.2\% &\cellcolor[HTML]{eae4fc}5.4\% &\cellcolor[HTML]{ddd3f9}8.7\% \\
Arab &(4) &\cellcolor[HTML]{fdfcff}0.7\% &\cellcolor[HTML]{f1ecfd}3.8\% &\cellcolor[HTML]{fefeff}0.4\% &\cellcolor[HTML]{fffeff}0.2\% \\
Jpan &(1) &\cellcolor[HTML]{faf8ff}1.4\% &\cellcolor[HTML]{fbfaff}1.1\% &\cellcolor[HTML]{faf8fe}1.4\% &\cellcolor[HTML]{fcf1ef}-1.0\% \\
Hani &(1) &\cellcolor[HTML]{fefefe}-0.0\% &\cellcolor[HTML]{fffeff}0.2\% &0.0\% &\cellcolor[HTML]{fefdfc}-0.1\% \\
Other &(20) &\cellcolor[HTML]{fbf9ff}1.2\% &\cellcolor[HTML]{f7f4fe}2.3\% &\cellcolor[HTML]{e8e1fb}5.9\% &\cellcolor[HTML]{fefdff}0.5\% \\
\bottomrule
\end{tabular}
\caption{Relative improvement (\%) of continuing tokenizer training over extending from a tokenizer trained from scratch (naive) in \textit{bytes per token}. We calculate the mean improvement across extension sizes 1k, 2k, 4k, 8k, 16k, 32k and 70 languages grouped by scripts.}\label{tab:70-lang-script}
\end{table}
\paragraph{Why is continued BPE training more effective?} In both vocabulary extension methods, each newly added token has merges leading to it from the tokens already in the tokenizer. However, this does not guarantee that these tokens will actually be produced during inference. To quantify this, we introduce the Self-Tokenization Test (see Section~\ref{sec:detecting-unreach-tokens}), which calculates \textit{tokens unreachable through merges}. \textbf{Extending from an independently trained tokenizer (naive) produces more unreachable tokens compared to the continued tokenizer training, which does not produce any} (see Table~\ref{tab:70-lang-avg}). This means that naive vocabulary extension has added tokens that will never be used in the merge process. 

\begin{table}[!tp]\centering
\addtolength{\tabcolsep}{-3.1pt}
\small

\begin{tabular}{lcccccccc}\toprule
&\multicolumn{3}{c}{\textbf{Llama-3}} & &\multicolumn{3}{c}{\textbf{Qwen-2.5}} \\\cmidrule{2-4}\cmidrule{6-8}
\textbf{Ext. size} &\textbf{ours} &\textbf{naive} &\textbf{WR} & &\textbf{ours} &\textbf{naive} &\textbf{WR} \\\midrule
+1000 &0.0\% &\cellcolor[HTML]{fdf1ee}4.0\% &85.7\% & &0.0\% &\cellcolor[HTML]{fcefeb}8.3\% &88.6\% \\
+2000 &0.1\% &\cellcolor[HTML]{fdf0ed}4.3\% &92.9\% & &0.0\% &\cellcolor[HTML]{fcece8}9.7\% &98.6\% \\
+4000 &0.1\% &\cellcolor[HTML]{fcefeb}4.6\% &98.6\% & &0.1\% &\cellcolor[HTML]{fbe9e4}11.0\% &100.0\% \\
+8000 &\cellcolor[HTML]{fffefe}0.5\% &\cellcolor[HTML]{fcede9}5.1\% &98.6\% & &0.4\% &\cellcolor[HTML]{fbe6e0}12.6\% &100.0\% \\
+16000 &\cellcolor[HTML]{fef8f7}2.1\% &\cellcolor[HTML]{fbe6e1}6.9\% &100.0\% & &\cellcolor[HTML]{fffbfb}2.0\% &\cellcolor[HTML]{fae0d9}15.3\% &100.0\% \\
+32000 &\cellcolor[HTML]{fbe6e0}7.1\% &\cellcolor[HTML]{f7d4cb}11.8\% &100.0\% & &\cellcolor[HTML]{fdf1ee}7.0\% &\cellcolor[HTML]{f7d4cb}20.9\% &100.0\% \\
\bottomrule
\end{tabular}
\caption{Unused added tokens in practice ($\downarrow$, average across 70 languages). Percentage of the added tokens that were not produced when tokenizing 10,000 target-language documents sampled from Fineweb-2 test set.}\label{tab:unused}

\end{table}

It is important to note that the Llama-3 and Mistral Nemo use merge skipping by default, so some of the tokens unreachable through merges will still be produced in practice.
To quantify the effect of this, we disable merge skipping for Llama-3 and Mistral Nemo and find that for the naive method, the compression drops 1.2\%--5.8\% and 0.5\%--4.1\% respectively on average across 70 languages  depending on the extension size (see Table~\ref{tab:70-lang-mergeskipping} in Appendix~\ref{sec:merge-skipping}). For our method, the difference is less than 0.1\% on average. This might explain why the difference between our method and the naive method is smaller for Llama-3 and Mistral Nemo, that use merge skipping, compared to Qwen-2.5 and Llama-2, which do not.

\begin{table*}[!htp]\centering
\small
\begin{tabular}{lrcccccccccc}\toprule
&\multirow{2.4}{*}{\textbf{$N_{\Delta}$}} &\multicolumn{5}{c}{\textbf{Continued tokenizer training (ours)}} &\multicolumn{5}{c}{\textbf{Naive extension}} \\\cmidrule(rl){3-7}\cmidrule(rl){8-12}
& &\textbf{ID} &\textbf{ID (L*)} &\textbf{Freq} &\textbf{Freq (L*)} &\textbf{Merge} &\textbf{ID} &\textbf{ID (L*)} &\textbf{Freq} &\textbf{Freq (L*)} &\textbf{Merge} \\\midrule
\multirow{8}{*}{\rotatebox[origin=c]{90}{\textbf{ET} (BPT)}} &\textbf{1000} &3.289 &3.289 &3.289 &3.289 &3.289 &3.164 &3.164 &3.164 &3.164 &3.164 \\
&\textbf{2000} &\cellcolor[HTML]{fbf9ff}3.526 &\cellcolor[HTML]{fbf9ff}3.526 &\cellcolor[HTML]{fbf9ff}3.527 &\cellcolor[HTML]{fbf9ff}3.527 &\cellcolor[HTML]{fbf9ff}3.527 &\cellcolor[HTML]{fbfaff}3.362 &\cellcolor[HTML]{fbfaff}3.362 &\cellcolor[HTML]{fbfaff}3.362 &\cellcolor[HTML]{fbfaff}3.362 &\cellcolor[HTML]{fbfaff}3.362 \\
&\textbf{4000} &\cellcolor[HTML]{f5f2fe}3.786 &\cellcolor[HTML]{f5f2fe}3.786 &\cellcolor[HTML]{f5f2fe}3.788 &\cellcolor[HTML]{f5f2fe}3.788 &\cellcolor[HTML]{f5f2fe}3.788 &\cellcolor[HTML]{f6f3fe}3.592 &\cellcolor[HTML]{f6f3fe}3.592 &\cellcolor[HTML]{f6f3fe}3.595 &\cellcolor[HTML]{f6f3fe}3.595 &\cellcolor[HTML]{f6f3fe}3.595 \\
&\textbf{8000} &\cellcolor[HTML]{eee9fc}4.109 &\cellcolor[HTML]{eee9fc}4.109 &\cellcolor[HTML]{eee9fc}4.110 &\cellcolor[HTML]{eee9fc}4.110 &\cellcolor[HTML]{eee9fc}4.110 &\cellcolor[HTML]{f0ebfd}3.870 &\cellcolor[HTML]{f0ebfd}3.870 &\cellcolor[HTML]{f0ebfd}3.874 &\cellcolor[HTML]{f0ebfd}3.874 &\cellcolor[HTML]{f0ebfd}3.874 \\
&\textbf{16000} &\cellcolor[HTML]{e7e0fb}4.451 &\cellcolor[HTML]{e7e0fb}4.452 &\cellcolor[HTML]{e7e0fb}4.456 &\cellcolor[HTML]{e7e0fb}4.456 &\cellcolor[HTML]{e7e0fb}4.456 &\cellcolor[HTML]{e9e2fb}4.202 &\cellcolor[HTML]{e9e2fb}4.202 &\cellcolor[HTML]{e9e2fb}4.208 &\cellcolor[HTML]{e9e2fb}4.208 &\cellcolor[HTML]{e9e2fb}4.208 \\
&\textbf{32000} &\cellcolor[HTML]{e0d6fa}4.829 &\cellcolor[HTML]{e0d6fa}4.829 &\cellcolor[HTML]{dfd6fa}4.837 &\cellcolor[HTML]{dfd6fa}4.837 &\cellcolor[HTML]{dfd6fa}4.837 &\cellcolor[HTML]{e1d8fa}4.552 &\cellcolor[HTML]{e1d8fa}4.552 &\cellcolor[HTML]{e1d8fa}4.567 &\cellcolor[HTML]{e1d8fa}4.567 &\cellcolor[HTML]{e1d8fa}4.566 \\
&\textbf{64000} &\cellcolor[HTML]{d9cdf8}5.164 &\cellcolor[HTML]{d9cdf8}5.164 &\cellcolor[HTML]{d8ccf8}5.197 &\cellcolor[HTML]{d8ccf8}5.198 &\cellcolor[HTML]{d8ccf8}5.198 &\cellcolor[HTML]{d9cef9}4.907 &\cellcolor[HTML]{d9cef9}4.907 &\cellcolor[HTML]{d9cdf8}4.943 &\cellcolor[HTML]{d9cdf8}4.943 &\cellcolor[HTML]{d9cdf8}4.943 \\
&\textbf{112000} &\cellcolor[HTML]{d4c7f8}5.381 &\cellcolor[HTML]{d4c7f8}5.381 &\cellcolor[HTML]{d3c6f7}5.407 &\cellcolor[HTML]{d4c7f8}5.386 &\cellcolor[HTML]{d4c7f8}5.386 &\cellcolor[HTML]{d4c7f8}5.147 &\cellcolor[HTML]{d4c7f8}5.147 &\cellcolor[HTML]{ddd3f9}4.732 &\cellcolor[HTML]{d4c7f8}5.154 &\cellcolor[HTML]{d3c6f7}5.174 \\
\midrule
\multirow{8}{*}{\rotatebox[origin=c]{90}{\textbf{EN} (BPT)}} &\textbf{1000} &\cellcolor[HTML]{d4c7f8}4.862 &\cellcolor[HTML]{d4c7f8}4.862 &\cellcolor[HTML]{d4c7f8}4.862 &\cellcolor[HTML]{d4c7f8}4.862 &\cellcolor[HTML]{d4c7f8}4.862 &\cellcolor[HTML]{d4c7f8}4.862 &\cellcolor[HTML]{d4c7f8}4.862 &\cellcolor[HTML]{d4c7f8}4.862 &\cellcolor[HTML]{d4c7f8}4.862 &\cellcolor[HTML]{d4c7f8}4.862 \\
&\textbf{2000} &\cellcolor[HTML]{d4c7f8}4.862 &\cellcolor[HTML]{d4c7f8}4.862 &\cellcolor[HTML]{d4c7f8}4.862 &\cellcolor[HTML]{d4c7f8}4.862 &\cellcolor[HTML]{d4c7f8}4.862 &\cellcolor[HTML]{d4c7f8}4.862 &\cellcolor[HTML]{d4c7f8}4.862 &\cellcolor[HTML]{d4c7f8}4.862 &\cellcolor[HTML]{d4c7f8}4.862 &\cellcolor[HTML]{d4c7f8}4.862 \\
&\textbf{4000} &\cellcolor[HTML]{d4c7f8}4.862 &\cellcolor[HTML]{d4c7f8}4.862 &\cellcolor[HTML]{d4c7f8}4.862 &\cellcolor[HTML]{d4c7f8}4.862 &\cellcolor[HTML]{d4c7f8}4.862 &\cellcolor[HTML]{d4c7f8}4.862 &\cellcolor[HTML]{d4c7f8}4.862 &\cellcolor[HTML]{d4c7f8}4.862 &\cellcolor[HTML]{d4c7f8}4.862 &\cellcolor[HTML]{d4c7f8}4.862 \\
&\textbf{8000} &\cellcolor[HTML]{d4c7f8}4.862 &\cellcolor[HTML]{d4c7f8}4.862 &\cellcolor[HTML]{d4c7f8}4.862 &\cellcolor[HTML]{d4c7f8}4.862 &\cellcolor[HTML]{d4c7f8}4.862 &\cellcolor[HTML]{d4c7f8}4.862 &\cellcolor[HTML]{d4c7f8}4.862 &\cellcolor[HTML]{d4c7f8}4.862 &\cellcolor[HTML]{d4c7f8}4.862 &\cellcolor[HTML]{d4c7f8}4.862 \\
&\textbf{16000} &\cellcolor[HTML]{d4c7f8}4.862 &\cellcolor[HTML]{d4c7f8}4.862 &\cellcolor[HTML]{d4c7f8}4.862 &\cellcolor[HTML]{d4c7f8}4.862 &\cellcolor[HTML]{d4c7f8}4.862 &\cellcolor[HTML]{d4c7f8}4.862 &\cellcolor[HTML]{d4c7f8}4.862 &\cellcolor[HTML]{d4c7f8}4.863 &\cellcolor[HTML]{d4c7f8}4.863 &\cellcolor[HTML]{d4c7f8}4.863 \\
&\textbf{32000} &\cellcolor[HTML]{d5c8f8}4.852 &\cellcolor[HTML]{d5c8f8}4.852 &\cellcolor[HTML]{d4c7f8}4.863 &\cellcolor[HTML]{d4c7f8}4.863 &\cellcolor[HTML]{d4c7f8}4.863 &\cellcolor[HTML]{d4c8f8}4.853 &\cellcolor[HTML]{d4c8f8}4.853 &\cellcolor[HTML]{d4c7f8}4.864 &\cellcolor[HTML]{d4c7f8}4.864 &\cellcolor[HTML]{d4c7f8}4.864 \\
&\textbf{64000} &\cellcolor[HTML]{dcd2f9}4.740 &\cellcolor[HTML]{dcd2f9}4.740 &\cellcolor[HTML]{d3c6f7}4.869 &\cellcolor[HTML]{d3c6f7}4.869 &\cellcolor[HTML]{d4c7f8}4.868 &\cellcolor[HTML]{dcd2f9}4.738 &\cellcolor[HTML]{dcd2f9}4.738 &\cellcolor[HTML]{d4c7f8}4.868 &\cellcolor[HTML]{d4c7f8}4.868 &\cellcolor[HTML]{d3c6f7}4.868 \\
&\textbf{112000} &4.219 &4.219 &\cellcolor[HTML]{f3effd}4.402 &\cellcolor[HTML]{f4f1fd}4.385 &\cellcolor[HTML]{f4f1fd}4.387 &4.206 &4.206 &\cellcolor[HTML]{f4f1fd}4.377 &\cellcolor[HTML]{f4f0fd}4.382 &\cellcolor[HTML]{f4f0fd}4.384 \\
\midrule
\multirow{8}{*}{\rotatebox[origin=c]{90}{\textbf{Unreach. tokens}}}&\textbf{1000} &\cellcolor[HTML]{fef7f6}556 &\cellcolor[HTML]{fef7f6}551 &\cellcolor[HTML]{fef8f6}541 &\cellcolor[HTML]{fef8f6}534 &\cellcolor[HTML]{fef8f6}534 &630 &625 &615 &608 &608 \\
&\textbf{2000} &\cellcolor[HTML]{fef8f6}536 &\cellcolor[HTML]{fef8f6}527 &\cellcolor[HTML]{fef8f6}509 &\cellcolor[HTML]{fef8f7}488 &\cellcolor[HTML]{fef8f7}504 &710 &701 &683 &662 &678 \\
&\textbf{4000} &\cellcolor[HTML]{fef8f7}483 &\cellcolor[HTML]{fef9f7}465 &\cellcolor[HTML]{fef9f8}446 &\cellcolor[HTML]{fefaf9}386 &\cellcolor[HTML]{fef9f8}445 &834 &817 &798 &738 &797 \\
&\textbf{8000} &\cellcolor[HTML]{fefaf9}371 &\cellcolor[HTML]{fffbfa}337 &\cellcolor[HTML]{fefaf9}371 &\cellcolor[HTML]{fffdfc}176 &\cellcolor[HTML]{fffbfa}328 &1273 &1239 &1272 &1077 &1230 \\
&\textbf{16000} &\cellcolor[HTML]{fffdfc}173 &\cellcolor[HTML]{fffdfd}136 &\cellcolor[HTML]{fdf1ee}980 &0 &\cellcolor[HTML]{fffefd}132 &\cellcolor[HTML]{fffefd}2196 &\cellcolor[HTML]{fffefd}2159 &\cellcolor[HTML]{fffdfc}3015 &\cellcolor[HTML]{fffefd}2035 &\cellcolor[HTML]{fffefd}2167 \\
&\textbf{32000} &0 &0 &\cellcolor[HTML]{f9ded6}2309 &0 &0 &\cellcolor[HTML]{fffbfa}4656 &\cellcolor[HTML]{fffbfa}4656 &\cellcolor[HTML]{fef8f7}7002 &\cellcolor[HTML]{fffbfa}4693 &\cellcolor[HTML]{fffbfa}4693 \\
&\textbf{64000} &0 &0 &\cellcolor[HTML]{f8d8cf}2711 &0 &0 &\cellcolor[HTML]{fef5f3}9581 &\cellcolor[HTML]{fef5f3}9581 &\cellcolor[HTML]{fdf2ef}12663 &\cellcolor[HTML]{fef5f3}9864 &\cellcolor[HTML]{fdf5f2}9868 \\
&\textbf{112000} &0 &0 &\cellcolor[HTML]{f7d4cb}2923 &0 &0 &\cellcolor[HTML]{fceeeb}15279 &\cellcolor[HTML]{fceeeb}15279 &\cellcolor[HTML]{f7d4cb}37637 &\cellcolor[HTML]{fcede9}16567 &\cellcolor[HTML]{fcede9}16813 \\
\bottomrule
\end{tabular}
\caption{\textbf{Interaction of pruning and extension} (Llama-3). Bytes-per-token (BPT) on \textsc{Flores} and unreachable tokens with different pruning and extension methods. In each experiment, we prune the tokenizer by $N_\Delta$ tokens using 50\%--50\% Estonian-English data mix, and then extend it back to the original size using Estonian-only data. *L indicates that leaf-based pruning was used.}\label{tab:pruning-bpt}
\end{table*}

In addition to calculating the number of tokens we can not reach through merges, we also tokenized the held-out target-language dataset using extended Llama-3 and Qwen-2.5 tokenizers to see which tokens are not produced in practice. On average, across 70 languages, our method yields a 4\%--13.9\% improvement in the utilization of the added token vocabulary in practice (see Table~\ref{tab:unused}). Furthermore, we observed more low-frequency tokens after naive extension compared to continued BPE training when visualizing Estonian token counts (see Appendix~\ref{sec:extension-histogram}).

\paragraph{Pruning.} When pruning and then extending the tokenizers\footnote{It is also possible to first extend and then prune; however, we did not observe any meaningful difference in performance depending on the order (see Appendix~\ref{sec:order-of-pruning}).} (see Table~\ref{tab:pruning-bpt}), we see that in terms of text compression, frequency-aware methods lead to better progress in added language (Estonian) and slower degradation in the base tokenizer language (English). In addition, leaf-based methods perform comparably or better than analogous naive implementations, as they account for the BPE structure. This is more evident in the number of unreachable tokens, where naive frequency-based pruning leads to a dramatic increase in unreachable tokens, breaking the BPE merge paths. Our results conclude that the leaf-based frequency pruning is the most effective approach. Merge-based pruning, our other proposed algorithm, performs comparably to the leaf-based pruning.

We provide a comparison of bilingual pruning without extension for nine languages in Appendix~\ref{sec:pruning-results-multi}, similarly demonstrating the benefits of leaf- and merge-based pruning methods.

\begin{table*}[!t]\centering
\scriptsize
\addtolength{\tabcolsep}{-3.4pt}

\begin{tabular}{@{}lcccccccccccccc@{}c@{}c@{}}\toprule
&\textbf{Compr.} & &\multicolumn{2}{c}{\textbf{FLORES} \tiny{(COMET)}} &\textbf{} &\multicolumn{2}{c}{\textbf{Winogrande} \tiny{(acc)}} &\textbf{} &\multicolumn{2}{c}{\textbf{Belebele} \tiny{(acc)}} &\textbf{} &\multicolumn{2}{c}{\textbf{SIB200} \tiny{(acc)}} &\textbf{} &\textbf{XCOPA} \tiny{(acc)} \\\cmidrule{4-5}\cmidrule{7-8}\cmidrule{10-11}\cmidrule{13-14}
\textbf{Model} &\textbf{ET \tiny{(EN)}} &\textbf{Toks.} &\textbf{EN-ET} &\textbf{ET-EN} &\textbf{} &\textbf{ ET} &\textbf{EN} &\textbf{} &\textbf{ET} &\textbf{EN} &\textbf{} &\textbf{ET} &\textbf{EN} &\textbf{} &\textbf{ET} \\\midrule
\cellcolor[HTML]{f3f3f3}\textbf{Llama-3.2-1B} &\cellcolor[HTML]{f3f3f3}\textbf{} &\cellcolor[HTML]{f3f3f3}\textbf{} &\cellcolor[HTML]{f3f3f3}\textbf{} &\cellcolor[HTML]{f3f3f3}\textbf{} &\cellcolor[HTML]{f3f3f3}\textbf{} &\cellcolor[HTML]{f3f3f3}\textbf{} &\cellcolor[HTML]{f3f3f3}\textbf{} &\cellcolor[HTML]{f3f3f3}\textbf{} &\cellcolor[HTML]{f3f3f3}\textbf{} &\cellcolor[HTML]{f3f3f3}\textbf{} &\cellcolor[HTML]{f3f3f3}\textbf{} &\cellcolor[HTML]{f3f3f3}\textbf{} &\cellcolor[HTML]{f3f3f3}\textbf{} &\cellcolor[HTML]{f3f3f3}\textbf{} &\cellcolor[HTML]{f3f3f3}\textbf{} \\
No training &2.64 \tiny{(4.86)} &0 &0.432 \tiny{± 0.008} &0.701 \tiny{± 0.007} & &50.0 \tiny{± 2.3} &61.2 \tiny{± 2.7} & &28.6 \tiny{± 3.0} &35.7 \tiny{± 3.1} & &71.6 \tiny{± 6.2} &77.0 \tiny{± 5.8} & &51.0 \tiny{± 4.4} \\
\textbf{Continued pretraining:} &\textbf{} &\textbf{} & & & & & & & & & & & & & \\
- default tokenizer &2.64 \tiny{(4.86)} &7.2B &0.788 \tiny{± 0.008} &0.803 \tiny{± 0.005} & &58.1 \tiny{± 2.3} &59.1 \tiny{± 2.7} & &28.0 \tiny{± 2.9} &26.9 \tiny{± 2.9} & &77.9 \tiny{± 5.7} &74.5 \tiny{± 6.0} & &63.6 \tiny{± 4.2} \\
- prune+ext (16k, naive) &4.21 \tiny{(4.86)} &5.4B &0.790 \tiny{± 0.008} &0.809 \tiny{± 0.005} & &57.4 \tiny{± 2.3} &58.8 \tiny{± 2.7} & &25.6 \tiny{± 2.9} &25.7 \tiny{± 2.9} & &73.0 \tiny{± 6.1} &73.0 \tiny{± 6.1} & &62.6 \tiny{± 4.2} \\
- prune+ext (16k, ours) &4.46 \tiny{(4.86)} &5.3B &0.796 \tiny{± 0.007} &0.803 \tiny{± 0.005} & &58.3 \tiny{± 2.3} &58.6 \tiny{± 2.7} & &27.3 \tiny{± 2.9} &27.1 \tiny{± 2.9} & &76.5 \tiny{± 5.8} &73.0 \tiny{± 6.1} & &63.0 \tiny{± 4.2} \\
& & & & & & & & & & & & & & & \\
\cellcolor[HTML]{f3f3f3}\textbf{Llama-3.2-3B} &\cellcolor[HTML]{f3f3f3}\textbf{} &\cellcolor[HTML]{f3f3f3}\textbf{} &\cellcolor[HTML]{f3f3f3}\textbf{} &\cellcolor[HTML]{f3f3f3}\textbf{} &\cellcolor[HTML]{f3f3f3}\textbf{} &\cellcolor[HTML]{f3f3f3}\textbf{} &\cellcolor[HTML]{f3f3f3}\textbf{} &\cellcolor[HTML]{f3f3f3}\textbf{} &\cellcolor[HTML]{f3f3f3}\textbf{} &\cellcolor[HTML]{f3f3f3}\textbf{} &\cellcolor[HTML]{f3f3f3}\textbf{} &\cellcolor[HTML]{f3f3f3}\textbf{} &\cellcolor[HTML]{f3f3f3}\textbf{} &\cellcolor[HTML]{f3f3f3}\textbf{} &\cellcolor[HTML]{f3f3f3}\textbf{} \\
No training &2.64 \tiny{(4.86)} &0 &0.635 \tiny{± 0.010} &0.796 \tiny{± 0.006} & &53.0 \tiny{± 2.3} &69.1 \tiny{± 2.5} & &44.1 \tiny{± 3.2} &74.2 \tiny{± 2.9} & &74.5 \tiny{± 6.0} &76.0 \tiny{± 5.9} & &55.4 \tiny{± 4.4} \\
\textbf{Continued pretraining:} &\textbf{} &\textbf{} & & & & & & & & & & & & & \\
- default tokenizer &2.64 \tiny{(4.86)} &7.2B &0.835 \tiny{± 0.006} &0.830 \tiny{± 0.005} & &65.1 \tiny{± 2.2} &68.2 \tiny{± 2.6} & &52.9 \tiny{± 3.3} &64.7 \tiny{± 3.1} & &80.9 \tiny{± 5.4} &77.0 \tiny{± 5.8} & &70.4 \tiny{± 4.0} \\
- prune+ext (16k, naive) &4.21 \tiny{(4.86)} &5.4B &0.818 \tiny{± 0.008} &0.831 \tiny{± 0.005} & &63.2 \tiny{± 2.2} &68.2 \tiny{± 2.6} & &51.1 \tiny{± 3.3} &66.0 \tiny{± 3.1} & &81.4 \tiny{± 5.4} &78.9 \tiny{± 5.6} & &69.8 \tiny{± 4.0} \\
- prune+ext (16k, ours) &4.46 \tiny{(4.86)} &5.3B &0.822 \tiny{± 0.007} &0.829 \tiny{± 0.005} & &63.1 \tiny{± 2.3} &68.4 \tiny{± 2.6} & &48.7 \tiny{± 3.3} &65.0 \tiny{± 3.1} & &78.4 \tiny{± 5.7} &77.9 \tiny{± 5.7} & &71.8 \tiny{± 3.9} \\
\bottomrule
\end{tabular}
\caption{Downstream task evaluation results on Llama-3 family of models after continued pre-training with English-Estonian data mix for different tokenizer setups (see Table~\ref{tab:downstream-tokenizers} for tokenizer metrics).}\label{tab:downstream-results}

\end{table*}

\subsection{Downstream evaluation}
\subsubsection{Continued pre-training}
Downstream task results for the continually pre-trained Llama-3.2 1B and 3B models in Table~\ref{tab:downstream-results} show no substantial difference between the naive and continued tokenizer training approaches. In these experiments, we first applied leaf-based frequency pruning and then extended the vocabulary to maintain a constant size. When comparing the original and modified tokenizers, the 1B model performs similarly in both cases, while the 3B model exhibits slightly lower scores on EN-ET machine translation and Belebele (ET) for the modified tokenizers, while having similar results on other benchmarks. Nevertheless, modifying the tokenizer reduced total training time by 26\% compared to the base tokenizer, owing to improved compression for Estonian (see Table~\ref{tab:downstream-tokenizers}). When looking at the embeddings, we observe signs of undertrained tokens for the naive method after continued pre-training compared to our method. A detailed analysis is provided in Appendix~\ref{sec:extension-histogram}.

\subsubsection{Model Compression via Pruning}
\label{sec:model-compression-pruning}
If we consider the scenario of using a model for a specific language or a domain, multilingual tokenizers typically include tokens for many languages that may not be required. We investigate how many of those tokens we can remove without affecting the downstream performance. Focusing on Estonian and English, we remove tokens from the vocabulary and the embeddings while aiming to preserve performance in these two languages.

\begin{figure*}[!t]
    \centering
    \includegraphics[width=\linewidth]{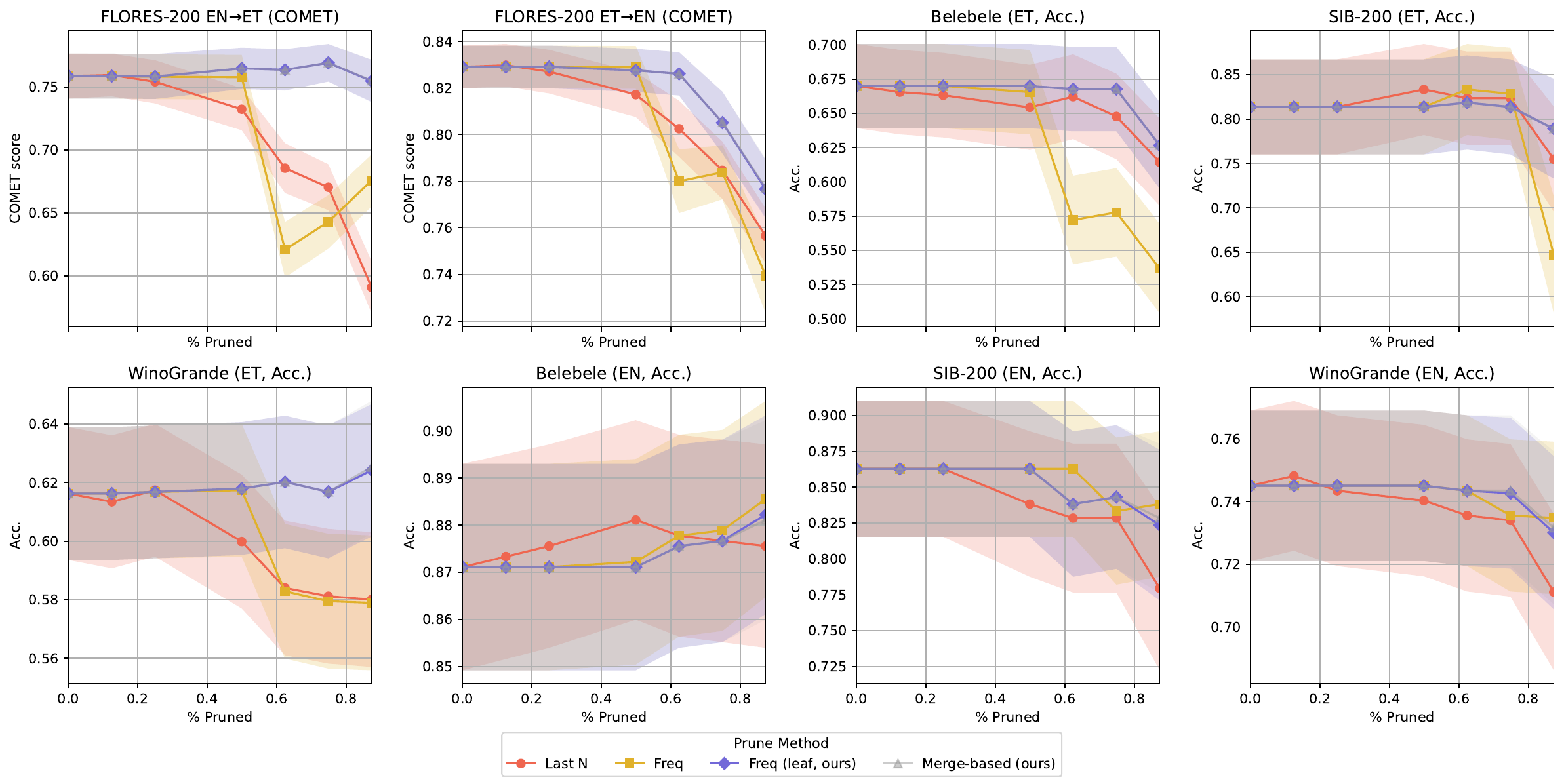}
    \caption{Llama-3.1-8B vocabulary pruned with different methods and evaluated on Estonian and English downstream tasks. The dataset for training the pruner consisted of 50-50 Estonian-English mix. The highlighted area shows the 95\% confidence interval acquired through bootstrap resampling. Note that \textit{Freq (leaf)} and \textit{Merge-based} method results overlap.}
    \label{fig:pruning-downstream}
\end{figure*}

\textbf{With our pruning methods, up to 62.5\% (80k/128k) of the vocabulary tokens can be removed without any noticeable deterioration in downstream performance} (see Figure~\ref{fig:pruning-downstream}). When comparing pruning strategies, leaf-based frequency pruning and merge-based pruning consistently allow for more extensive token removal than last-N pruning and naive frequency pruning. The latter two methods exhibit performance degradation earlier, whereas our approaches maintain robustness under higher pruning ratios. We also observe that machine translation (generative) is more sensitive to vocabulary pruning than the discriminative tasks, where the amount of tokens that can be removed without loss in performance is even higher than 62.5\%. We also observe similar results for German-English pruning (see Figure~\ref{fig:pruning_downstream_de} in Appendix~\ref{sec:pruning-results-multi}).

\section{Discussion}

Our method of continued tokenizer training yields better intrinsic tokenizer performance than the commonly utilized naive extension method. More importantly, unlike naive extension, it does not harm the BPE structure and leaves no unreachable tokens. Unreachable tokens introduced by the naive method are only partially addressed by merge skipping, as merge skipping uses only the tokens that constitute complete pre-tokens.

Leaf-based pruning enables structure-aware tokenizer trimming, which can be utilized in conjunction with the tokenizer extension or independently to reduce model size, a factor that is now largely dependent on embedding parameters, particularly for small models. Furthermore, introducing frequencies into this pruning approach makes it possible to remove the vocabulary needed the least. For example, it is possible to combine a representative sample of all languages or domains required in the fine-tuned model, not only the ones directly participating in the tokenizer extension.
%

\section{Conclusion}
We presented methods for controlled vocabulary modification of pre-trained language models through improved tokenizer extension and pruning. Our proposed continued BPE training extends an existing tokenizer directly on domain-specific data, avoiding the inefficiencies of appending tokens from an independently trained tokenizer. Experiments across multiple languages and model families demonstrate that this approach enhances tokenization efficiency and more effectively utilizes the extended vocabulary. In parallel, our leaf-based pruning algorithm enables safe vocabulary reduction by removing redundant tokens without degrading model performance. Together, these techniques provide a practical framework for adapting tokenizers to new domains and languages while maintaining model compactness and quality. To help facilitate further research on vocabulary adaptation, we release an open-source toolkit implementing our methods, while preserving compatibility with the Hugging Face format.

\section*{Limitations}

While our method demonstrates improved tokenization efficiency through continued BPE training, several limitations remain.

Our experiments were conducted in a bilingual setting, where a single language was added to an existing tokenizer. Although this setup is typical in low-resource adaptation, more complex multilingual scenarios remain unexplored and warrant future investigation.

The current approach is also restricted to BPE-based tokenizers, which are still the predominant choice in large language models. Extending the method to other segmentation algorithms could provide additional insights.

Although our evaluation focuses on large language models, the method itself is not inherently limited to them. It could be applied to any pre-trained model that relies on a BPE tokenizer. However, other architectures, such as encoder-decoder NMT models or encoder-only language models, were not included in our experiments.

We deliberately limit the scope of this work to tokenizer-level modification and do not explore how to achieve the best model-level downstream performance. This question likely depends on multiple interacting factors, such as dataset size, model capacity, and embedding initialization strategies, and thus requires a more extensive investigation. Consequently, we do not address when a tokenizer should be replaced entirely or how much tokenizer extension is optimal from the perspective of downstream performance.

Finally, the extrinsic evaluation on downstream tasks was, for the most part, limited to a single language and model, which may affect the generalizability of our findings to other settings.

\section*{Acknowledgments}
This work was supported by the National Program for Estonian Language Technology Program (project EKTB104) funded by the Estonian Ministry
of Education and Research. The authors also acknowledge the HPC resource allocation by Erlangen National High-Performance Computing Center (NHR@FAU) of the Friedrich-Alexander-Universität Erlangen-Nürnberg (FAU) in a joint project with the Center for Artificial Intelligence (CAIRO), THWS.

\bibliography{custom}

\appendix
\label{sec:appendix}

\section{Merge-based pruning}
\label{sec:merge-based-pruning}
\begin{algorithm}[!h]
\small
\DontPrintSemicolon
\SetAlgoLined
\KwInput{%
\\
  \Indp\var{vocab}: mapping from token $\to$ vocabulary index \\
  \var{tok\_counts}: token frequencies in tokenized corpus \\
  \var{merge\_counts}: frequencies of merges performed when tokenizing the corpus
}
\KwOutput{%
\\
\Indp
  \var{prune\_order}: list of tokens to prune in the order of pruning.
}
\;
\kw{initialize} \var{counts} $\gets$ \var{tok\_counts}\;

\vspace{0.3em}
\For{$(t_1, t_2)$, $n$ in \var{merge\_counts}}{
    \var{counts}[$t_1$] $\gets$ \var{counts}[$t_1$] + $n$\;
    \var{counts}[$t_2$] $\gets$ \var{counts}[$t_2$] + $n$\;
}

\vspace{0.3em}
\textbf{return} \kw{sort} tokens by ascending \var{counts}, break ties by descending length of the token and finally \var{vocab} index\;

\caption{Merge-Based Pruning (MBP)}\label{alg:mpb}
\end{algorithm}
As an alternative to the frequency based pruning with leaf ordering, we also developed the merge-based pruning (see Algorithm~\ref{alg:mpb}), which achieves similar results. The main idea is to count the token frequency throuout the mergining process. So, in practice this means counting the final token frequency and any time it appears in a merge, creating another token. Consequently, a vocabulary token will have a count of $n$, while any token it was merged from will have a count of at least $n$. This follows from the fact that standard BPE defines a single deterministic merge tree for each token, and tokens appearing earlier in such trees often participate in multiple merge trees. This means that if we sort by that count, breaking ties by token length (when a token and its predecessor in the merge tree have equal count) and token index, only the leaf nodes will be removed.

Since the result is very similar we decided to focus on the leaf-frequency pruning in the main part of the paper.

\section{Additional results}
\subsection{Training dataset size}
\label{sec:budget}
\begin{figure}[!ht]
    \centering
    \includegraphics[width=\linewidth]{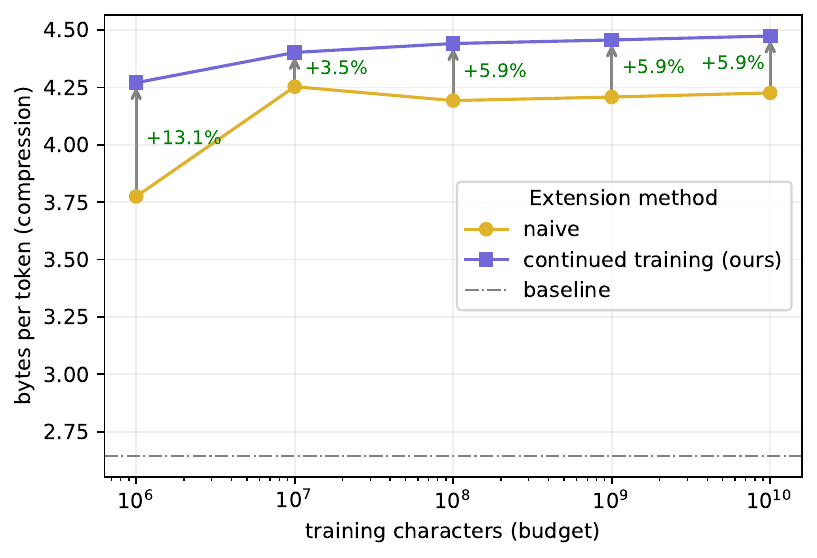}
    \caption{Extension training budget in characters vs compression of Estonian text when adding 16,000 Estonian tokens to the Llama-3 tokenizer.}
    \label{fig:budget}
\end{figure}

The effect of the training set diminishes at higher sizes. We see minimal difference between 1B and 10B character budget in scores when extending the tokenizer (see Figure~\ref{fig:budget}). Extending a tokenizer using continued training with 10M character training set yields a bigger effect than extending from a new tokenizer (naive) using 10B character training set --  the choice of extension method is more important than the training budget at this scale.

\subsection{Analysis of undertrained tokens}
\label{sec:extension-histogram}

\begin{figure}[h!]
    \centering
    \includegraphics[width=0.95\linewidth]{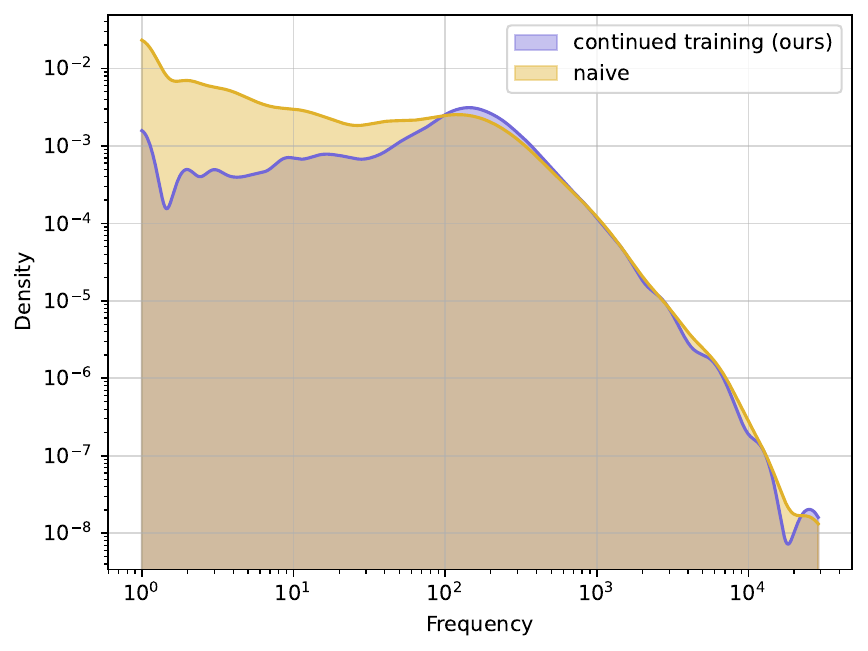}
    \caption{Token frequency distribution in the added tokens ($n=8000$) in Llama-3 tokenizer on the held-out set of Estonian Fineweb-2 ($10000$ documents).}
    \label{fig:ext-hist}
\end{figure}

Frequency distribution of the added tokens is visualized in Figure~\ref{fig:ext-hist}. We observe that the naive method yields more low-frequency tokens than our continued BPE training method.

\begin{figure}[h!]
    \centering
    \includegraphics[width=0.95\linewidth]{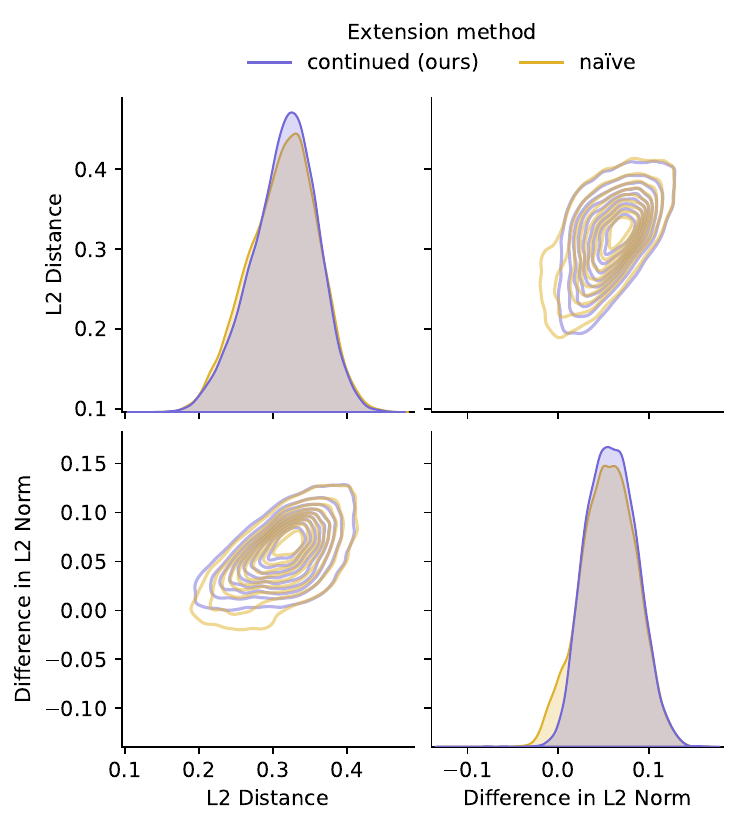}
    \caption{Comparison of input/output embeddings (tied) of added tokens at FVT initialization and after continued pre-training (L2 distance between them and the difference in L2 norms). We extend Llama-3.2-3B with Estonian tokens and continually pre-train on Estonian-English mix, comparing naive tokenizer extension and continued BPE training.}
    \label{fig:embeddings}
\end{figure}

To better understand how this affects the learning of new tokens, we analyze the changes in their embeddings during continued pre-training. Since the added tokens are initialized with FVT and updated during training, undertraining is not always immediately visible. To investigate this, we measure both the L2 distance of embeddings from their initialized values and the change in their L2 norms. Figure~\ref{fig:embeddings} shows that the naive extension method produces noticeably more undertrained tokens, reflected in smaller L2 changes and a stronger weight-decay effect that drives their embeddings toward zero (visible as a longer tail of negative L2 norm differences). This interpretation is further supported by Table~\ref{tab:unused}, which shows lower vocabulary utilization for the naive method.

\subsection{Extension effect on English}
\label{sec:effect-on-english}

Table~\ref{tab:effect-on-eng} demonstrates that the target-language extension both with continued BPE training (ours) and the naive method has almost no effect on the tokenization of English sentences.

\begin{table}[!htp]\centering
\small
\begin{tabular}{@{}lccccc@{}}\toprule
\textbf{} &\multicolumn{2}{c}{\textbf{Lev. dist. $\downarrow$}} &\multicolumn{2}{c}{\textbf{EM (\%) $\uparrow$}} \\\cmidrule(lr){2-3}\cmidrule(lr){4-5}
\textbf{Ext. size} &\textbf{ours} &\textbf{naive} &\textbf{ours} &\textbf{naive} \\\midrule
1000 &0.001 \tiny{$<$0.010} &0.001 \tiny{$<$0.010}  &100.0 &99.9 \\
2000 &0.002 \tiny{$<$0.012} &0.002 \tiny{$<$0.013}  &99.9 &99.9 \\
4000 &0.003 \tiny{$<$0.022} &0.004 \tiny{$<$0.022}  &99.9 &99.8 \\
8000 &0.009 \tiny{$<$0.036} &0.010 \tiny{$<$0.036}  &99.6 &99.6 \\
16000 &0.023 \tiny{$<$0.106} &0.024 \tiny{$<$0.098}  &99.1 &99.0 \\
32000 &0.057\tiny{ $<$0.198} &0.057 \tiny{$<$0.220}  &97.7 &97.6 \\
\bottomrule
\end{tabular}
\caption{Mean sentence-averaged Levenstein distance between the base tokenizer (Llama-3) and extended tokenizer tokenization of \textbf{English} FLORES sentences and the amount of sentences (\%) where the tokenizations are an exact match (EM). We report the mean over 70 languages and additionally the worst-case language Lev. dist ($<\max$).
}\label{tab:effect-on-eng}
\end{table}

\subsection{The effect of merge skipping}
\label{sec:merge-skipping}
We note that merge skipping contributes significantly to the compression. We look at models that use merge skipping by default (Llama-3 and Mistral Nemo) and disable merge skipping to see how much the performance drops. In Table~\ref{tab:70-lang-mergeskipping}, see that for continued BPE training, disabling merge skipping has no significant effect, while for the naive method, the drop in tokenization efficiency is notable. This is because merge skipping allows reaching tokens that are unreachable by merges if those tokens are equal to a pre-token, bypassing merges. As we previously noted, the naive method produces many of those unreachable tokens.

\begin{table}[!htp]\centering
\small
\begin{tabular}{lcccccc}\toprule
&\multicolumn{2}{c}{\textbf{Llama-3}} & &\multicolumn{2}{c}{\textbf{Mistral Nemo}} \\\cmidrule{2-3}\cmidrule{5-6}
&\textbf{ours} &\textbf{naive} &\textbf{} &\textbf{ours} &\textbf{naive} \\\midrule
+1000 &\cellcolor[HTML]{fefefe}-0.1\% &\cellcolor[HTML]{fdf5f3}-1.2\% &\textbf{} &0.0\% &\cellcolor[HTML]{fefbfa}-0.5\% \\
+2000 &\cellcolor[HTML]{fefefe}-0.1\% &\cellcolor[HTML]{fcf0ed}-1.9\% &\textbf{} &0.0\% &\cellcolor[HTML]{fdf8f6}-0.9\% \\
+4000 &\cellcolor[HTML]{fefefe}-0.1\% &\cellcolor[HTML]{fbeae5}-2.8\% &\textbf{} &0.0\% &\cellcolor[HTML]{fcf3f0}-1.6\% \\
+8000 &\cellcolor[HTML]{fefefe}-0.1\% &\cellcolor[HTML]{f9e2dc}-3.8\% &\textbf{} &0.0\% &\cellcolor[HTML]{fbede9}-2.4\% \\
+16000 &\cellcolor[HTML]{fefefe}-0.1\% &\cellcolor[HTML]{f8dad3}-4.9\% &\textbf{} &0.0\% &\cellcolor[HTML]{fae6e1}-3.3\% \\
+32000 &\cellcolor[HTML]{fefefe}-0.1\% &\cellcolor[HTML]{f7d4cb}-5.8\% &\textbf{} &0.0\% &\cellcolor[HTML]{f9e0d9}-4.1\% \\
\bottomrule
\end{tabular}
\caption{Change in compression (bytes per token) after disabling merge skipping depending on the tokenizer extension method. We report average change in performance across 70 languages for two tokenizers that use merge skipping.}\label{tab:70-lang-mergeskipping}

\end{table}

\subsection{Naive extension methods}
\label{sec:naive-extension-methods}
There are multiple ways of adding tokens from an auxiliary independently trained tokenizer to an existing tokenizer (naive method).

\begin{figure}[h!]
    \centering
    \includegraphics[width=0.95\linewidth]{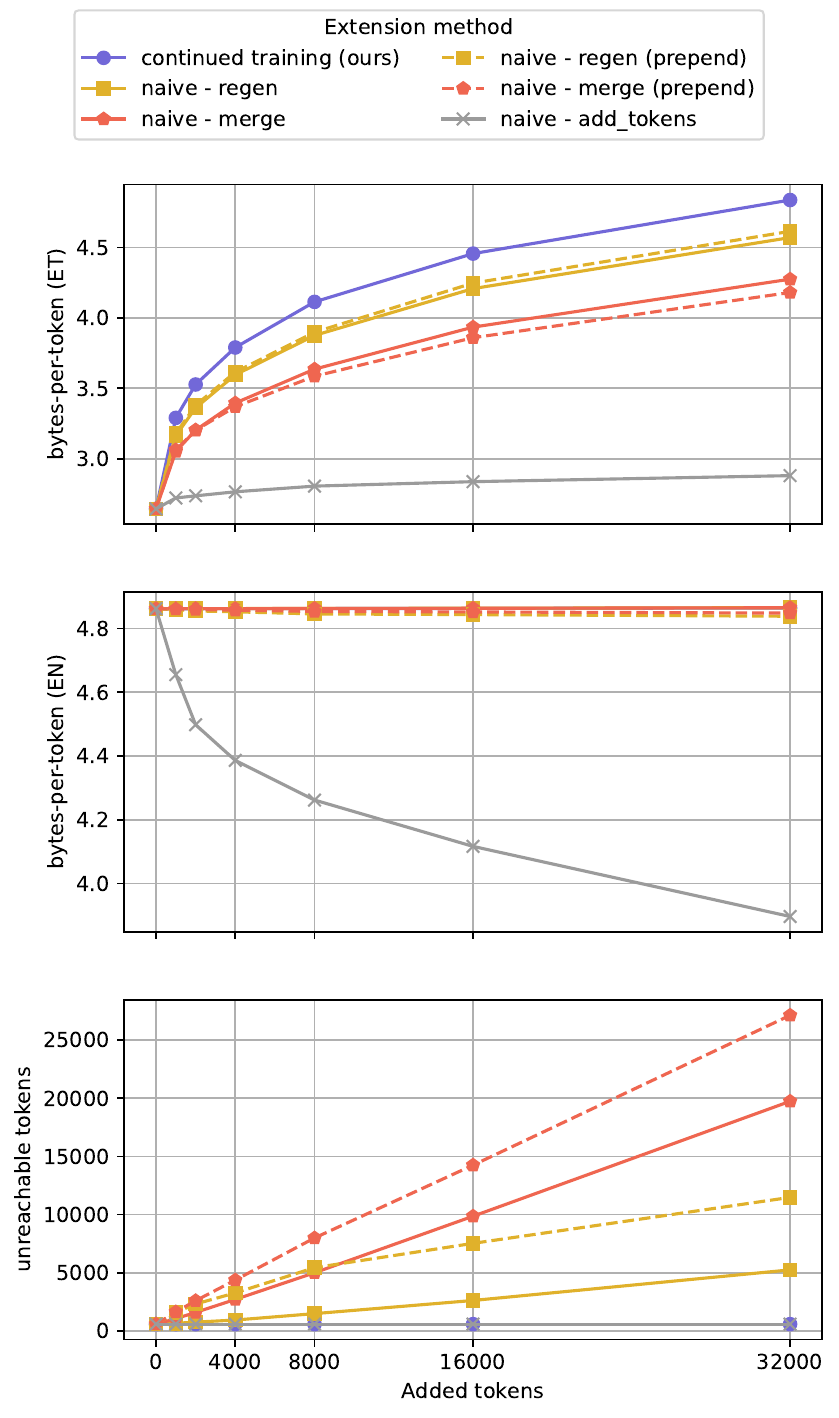}
    \caption{Comparison of \textbf{compression} (\textit{bytes per tokens}$\uparrow$) of Estonian (target language) and English, and \textbf{unreachable tokens}$\downarrow$ given the number of added tokens for various methods of adding new tokens for Llama-3.}
    \label{fig:naive-ext-methods}
\end{figure}

For tokenizers that do not define merge lists explicitly (e.g. SentencePiece), it is common to just append the vocabulary \cite{csaki-etal-2024-sambalingo, cui2024efficienteffectivetextencoding, tejaswi-etal-2024-exploring}. For Hugging Face (HF) transformers tokenizer implementations, it is more complicated because merges are performed according to a merge list, so in addition to adding tokens, we also need to handle the merges. Some works have just appended the merge list of a new tokenizer \cite{csaki2023efficientlyadaptingpretrainedlanguage, tejaswi-etal-2024-exploring}, however, we show later that this is not optimal. 

We took the approach of generating merges for the added vocabulary (\textit{regen}) by checking which tokens can create the added token and then ordering the merge rules by the priority/index. Our implementation is based on the way that HF transformers converts SentencePiece tokenizers to their format, which requires creating merge lists. This is also similar to what happens when tokens are added to a SentencePiece tokenizer, since there are no merge lists that are merged together -- the added tokens are produced through merging according to their priority.

In Figure~\ref{fig:naive-ext-methods}, we observe that the approach of generating merge rules for the added tokens (\textit{regen}) works better than appending merge rules (\textit{merge}). We also compared against using the \texttt{add\_tokens} method of HF transformers, however we find it has the lowest performance.

There are also works that have prepended merges \cite{csaki2023efficientlyadaptingpretrainedlanguage}. When used with the regen method, it does result in higher scores, however we see that it affects the English tokenization and results in more unreachable tokens, so we do not use it.

Thus we decide to use the \textit{regen} method, since we try to use the strongest baseline to compare our method against.

\subsection{Order of pruning/extension}
Table~\ref{tab:prune-ext-order} shows that the difference between pruning and extending (in that order) vs extending and pruning is negligible in terms of compression.

\label{sec:order-of-pruning}
\begin{table}[!htp]\centering
\small
\begin{tabular}{ccccc}\toprule
&\multicolumn{2}{c}{\textbf{Prune-Extend}} &\multicolumn{2}{c}{\textbf{Extend-prune}} \\\cmidrule(r){2-3}\cmidrule(l){4-5}
$N_\Delta$ &ET &EN &ET &EN \\\midrule
16000 &4.441 &4.863 &4.441 &4.863 \\
32000 &4.441 &4.863 &4.441 &4.863 \\
64000 &4.440&4.861 &4.439 &4.861 \\
96000 &4.402 &4.671 &4.401 &4.671 \\
\bottomrule
\end{tabular}
\caption{Order of pruning and extension. Compression (bytes per token) for Llama-3 extension (Estonian) and pruning (Estonian-English). $N_\Delta$ is the number of tokens pruned and then extended (or vice versa).
}\label{tab:prune-ext-order}
\end{table}

\section{Continued pre-training details}
\label{sec:downstream-tokenizers}
We provide details for the modified Llama-3 tokenizers in the continued pertaining experiments in Table~\ref{tab:downstream-tokenizers}.

The hyperparamters for continued pre-training using Hugging Face \texttt{transformers} \cite{wolf-etal-2020-transformers} are in Table~\ref{tab:cpt-hyperparams}. The training was conducted on the LUMI supercomputer using 8 nodes, each consisting of 4 AMD MI250x GPUs (acting as 8 compute units). The continued pre-training experiments took 2311 GPU-hours (180--612 GPU-hours per experiment).

\begin{table}[!t]\centering
\scriptsize
\addtolength{\tabcolsep}{-3pt}
\begin{tabular}{@{}lrrccccc@{}}\toprule
& & &\multicolumn{2}{c}{\textbf{Compr.}} &\textbf{CPT} & \\\cmidrule{4-5}
\textbf{Tokenizer} &\textbf{|vocab|} &\textbf{unreach} &\textbf{ET} &\textbf{EN} &\textbf{tokens} &\textbf{\% ET} \\\midrule
default &128k &588 &2.64 &4.86 &7.2B &63.9\% \\
prune+ext (16k) &128k &0 &4.46 &4.86 &5.3B &50.6\% \\
prune+ext (16k, naive) &128k &2033 &4.21 &4.86 &5.4B &52.0\% \\
\bottomrule
\end{tabular}
\caption{Continued pertaining (CPT) tokenizer overview.}\label{tab:downstream-tokenizers}

\end{table}

\begin{table}[!htp]\centering

\begin{tabular}{lrr}\toprule
Hyperparameter &Value \\\midrule
Learning rate &4e-4 (1B) / 1e-4 (3B)\\
Optimizer &AdamW \\
Adam $\epsilon$ &1e-8 \\
Adam $\beta_1, \beta_2$ &0.9, 0.95  \\
Sequence length &4096 \\
Weight decay &0.1 \\
Scheduler &cosine decayed to 10\% \\
Warmup ratio &10\% \\
FSDP Strategy &\texttt{SHARD\_GRAD\_OP} \\
GPUs &64 \\
Precision &\texttt{bfloat16} \\
Batch size &1024 \\
Batch size (tokens) &4194304 \\
\bottomrule
\end{tabular}
\caption{Continued pre-training hyperparameters.}\label{tab:cpt-hyperparams}

\end{table}

\section{Evaluation details}
\label{sec:evaluation-details}

We provide additional evaluation dataset details in Table~\ref{tab:eval-data}. The evaluation prompt setup is detailed in Figure~\ref{fig:eval-prompts}. For discriminative benchmarks, we report accuracy and 95\% confidence interval calculated from the standard error reported by \texttt{lm-eval-harness}. For FLORES-200 (MT) evaluation we calculate the confidence intervals using bootstrap resampling.

\begin{figure}[!tbp]
  \centering
   \newtcolorbox{promptbox}{
  colback=white,
  colframe=black,
  arc=4pt,
  boxrule=0.5pt,
  left=6pt,
  right=6pt,
}
\begin{minipage}{\linewidth}
\begin{promptbox}
\begin{small}
\textbf{FLORES} (\textit{generative})\\
\texttt{Translate the text from \{src\_lang\} to \{tgt\_lang\}.
\textbackslash{}n\\
\textbackslash{}n\\
English: \{src\_sent\}\textbackslash{}n\\
Estonian:\\
}
\rule{\textwidth}{0.1pt} \\
\textbf{Belebele} (\textit{log-likelihood})\\
\texttt{
P: \{passage\}\textbackslash{}n\\
Q: \{question\}\textbackslash{}n\\
A: \{answer1\}\textbackslash{}n\\
B: \{answer2\}\textbackslash{}n\\
C: \{answer3\}\textbackslash{}n\\
D: \{answer4\}\textbackslash{}n\\
Answer: \textcolor{gray}{\{correct letter\}}}
\\
\rule{\textwidth}{0.1pt} \\

\textbf{SIB-200} (\textit{log-likelihood})\\
\texttt{Topic Classification: science/technology, travel, politics, sports, health, entertainment, geography.\textbackslash{}n
\\\textbackslash{}n
\\
The topic of "[\{text\}]" is: \textcolor{gray}{\{topic\}}\\
}
\rule{\textwidth}{0.1pt} \\
\textbf{Winogrande} (\textit{log-likelihood})\\
\\
\texttt{\{pre\_blank\} \{choice\_1/choice\_2\} \textcolor{gray}{\{post\_blank\}}}

\rule{\textwidth}{0.1pt} \\

\textbf{X-COPA} (\textit{log-likelihood})\\
\texttt{\{premise\} \{connector\} \textcolor{gray}{\{choice\_1/choice\_2\}}}
\\

\end{small}
\end{promptbox}
\end{minipage}
  \caption{Prompts used for evaluation with \texttt{lm-eval-harness} \cite{eval-harness}.}\label{fig:eval-prompts}
\end{figure}

\section{Fast Vocabulary Transfer (FVT)}
\label{sec:fvt}
 To formally define Fast Vocabulary Transfer \cite[FVT][]{gee-etal-2022-fast}, let $V$ and $tok$ denote the vocabulary and tokenization function of the original tokenizer, and let $E(t)$ denote the embedding of token $t$. For a new tokenizer with vocabulary $V'$, the embedding $E'(t)$ for each $t \in V'$ is defined as:
\begin{equation}
E'(t) =
\begin{dcases}
  E(t) & \text{if } t \in V, \\
  \frac{\sum_{t_i \in tok(t)} E(t_i)}{|tok(t)|} & \text{otherwise}.
\end{dcases}
\end{equation}

\onecolumn
\begin{table*}[!htp]\centering
\small
\addtolength{\tabcolsep}{-2.5pt}
\begin{tabular}{llrrc@{}r}\toprule
\textbf{Dataset} &\textbf{} &\textbf{Size} &\textbf{Split} &\textbf{n-shots} & \textbf{License} \\\midrule
Winogrande & \cite{sakaguchi2021winogrande} &1267 &valid &0 & CC-BY \\
Winogrande ET & \cite{ojastu2025estonianwinogrande}&1767 &test &0 & Apache 2.0\\
XCOPA & \cite{ponti-etal-2020-xcopa} &600 &test &5 & CC-BY-4.0 \\
SIB200 & \cite{adelani-etal-2024-sib} &204 &test &5 & CC-BY-SA-4.0\\
FLORES200 & \cite{nllb2022} &1012 &devtest &5 & CC-BY-SA-4.0 \\
Belebele & \cite{bandarkar-etal-2024-belebele}&900 &test &5 & CC-BY-SA-4.0 \\
\bottomrule
\end{tabular}
\caption{Evaluation datasets. \textit{n-shot} is the number of in-context few-shot examples used for prompting during evaluation.}\label{tab:eval-data}
\end{table*}

\section{Languages in training data}
\label{sec:language-overview}

We provide the overview of the languages in the training dataset in Table~\ref{tab:70-lang-overview}.

\begin{table*}[!h]\centering
\scriptsize
\begin{tabular}{lllrllllr}\toprule
\textbf{Language} &\textbf{Code} &\textbf{Family} &\textbf{Bytes} &\textbf{} &\textbf{Language} &\textbf{Code} &\textbf{Family} &\textbf{Bytes} \\\cmidrule{1-4}\cmidrule{6-9}
Russian &rus\_Cyrl &Indo-European &6.4T & &Standard Estonian &ekk\_Latn &Uralic &43.8B \\
Mandarin Chinese &cmn\_Hani &Sino-Tibetan &2.7T & &Croatian &hrv\_Latn &Indo-European &38.6B \\
German &deu\_Latn &Indo-European &1.7T & &Standard Latvian &lvs\_Latn &Indo-European &35.8B \\
Japanese &jpn\_Jpan &Japonic &1.7T & &Standard Malay &zsm\_Latn &Austronesian &34.3B \\
Spanish &spa\_Latn &Indo-European &1.4T & &North Azerbaijani &azj\_Latn &Turkic &28.9B \\
French &fra\_Latn &Indo-European &1.2T & &Tamil &tam\_Taml &Dravidian &39.7B \\
Italian &ita\_Latn &Indo-European &793.8B & &Serbian &srp\_Cyrl &Indo-European &28.8B \\
Portuguese &por\_Latn &Indo-European &611.2B & &Tosk Albanian &als\_Latn &Indo-European &19.5B \\
Polish &pol\_Latn &Indo-European &463.9B & &Urdu &urd\_Arab &Indo-European &21.4B \\
Dutch &nld\_Latn &Indo-European &426.8B & &Kazakh &kaz\_Cyrl &Turkic &22.2B \\
Indonesian &ind\_Latn &Austronesian &374.4B & &Georgian &kat\_Geor &Kartvelian &27.1B \\
Turkish &tur\_Latn &Turkic &305.5B & &Nepali (individual language) &npi\_Deva &Indo-European &27.0B \\
Vietnamese &vie\_Latn &Austro-Asiatic &343.4B & &Marathi &mar\_Deva &Indo-European &24.2B \\
Czech &ces\_Latn &Indo-European &221.5B & &Malayalam &mal\_Mlym &Dravidian &23.9B \\
Standard Arabic &arb\_Arab &Afro-Asiatic &315.2B & &Macedonian &mkd\_Cyrl &Indo-European &16.1B \\
Korean &kor\_Hang &Koreanic &229.2B & &Icelandic &isl\_Latn &Indo-European &11.0B \\
Persian &fas\_Arab &Indo-European &327.1B & &Belarusian &bel\_Cyrl &Indo-European &12.3B \\
Hungarian &hun\_Latn &Uralic &214.4B & &Telugu &tel\_Telu &Dravidian &15.5B \\
Swedish &swe\_Latn &Indo-European &217.9B & &Afrikaans &afr\_Latn &Indo-European &8.3B \\
Romanian &ron\_Latn &Indo-European &199.9B & &Kannada &kan\_Knda &Dravidian &13.9B \\
Ukrainian &ukr\_Cyrl &Indo-European &273.7B & &Gujarati &guj\_Gujr &Indo-European &12.6B \\
Norwegian Bokmål &nob\_Latn &Indo-European &184.7B & &Galician &glg\_Latn &Indo-European &6.9B \\
Modern Greek (1453-) &ell\_Grek &Indo-European &238.4B & &Burmese &mya\_Mymr &Sino-Tibetan &13.3B \\
Thai &tha\_Thai &Kra-Dai &299.2B & &Moroccan Arabic &ary\_Arab &Afro-Asiatic &8.3B \\
Danish &dan\_Latn &Indo-European &161.8B & &Halh Mongolian &khk\_Cyrl &Mongolic &9.1B \\
Finnish &fin\_Latn &Uralic &153.6B & &Armenian &hye\_Armn &Indo-European &7.7B \\
Bulgarian &bul\_Cyrl &Indo-European &156.5B & &Khmer &khm\_Khmr &Austro-Asiatic &9.3B \\
Slovak &slk\_Latn &Indo-European &91.7B & &Northern Uzbek &uzn\_Latn &Turkic &4.8B \\
Hindi &hin\_Deva &Indo-European &129.9B & &Basque &eus\_Latn &Language isolate &4.6B \\
Lithuanian &lit\_Latn &Indo-European &60.7B & &Sinhala &sin\_Sinh &Indo-European &7.6B \\
Bosnian &bos\_Latn &Indo-European &52.8B & &Panjabi &pan\_Guru &Indo-European &6.1B \\
Hebrew &heb\_Hebr &Afro-Asiatic &73.8B & &Kirghiz &kir\_Cyrl &Turkic &4.7B \\
Bengali &ben\_Beng &Indo-European &93.5B & &Swahili (individual language) &swh\_Latn &Niger-Congo &3.3B \\
Slovenian &slv\_Latn &Indo-European &44.9B & &Norwegian Nynorsk &nno\_Latn &Indo-European &2.9B \\
Catalan &cat\_Latn &Indo-European &43.3B & &Odia &ory\_Orya &Indo-European &5.3B \\
\bottomrule
\end{tabular}
\caption{Overview of the 70 languages used. We report the number of UTF-8 bytes in Fineweb-2 \cite{penedo2025fineweb}.}\label{tab:70-lang-overview}

\end{table*}

\newpage
\section{Full results}
\subsection{Extended tokenizer pruning results}
\label{sec:pruning-results-multi}
\begin{figure*}[!hp]
    \centering
    \includegraphics[width=0.95\linewidth]{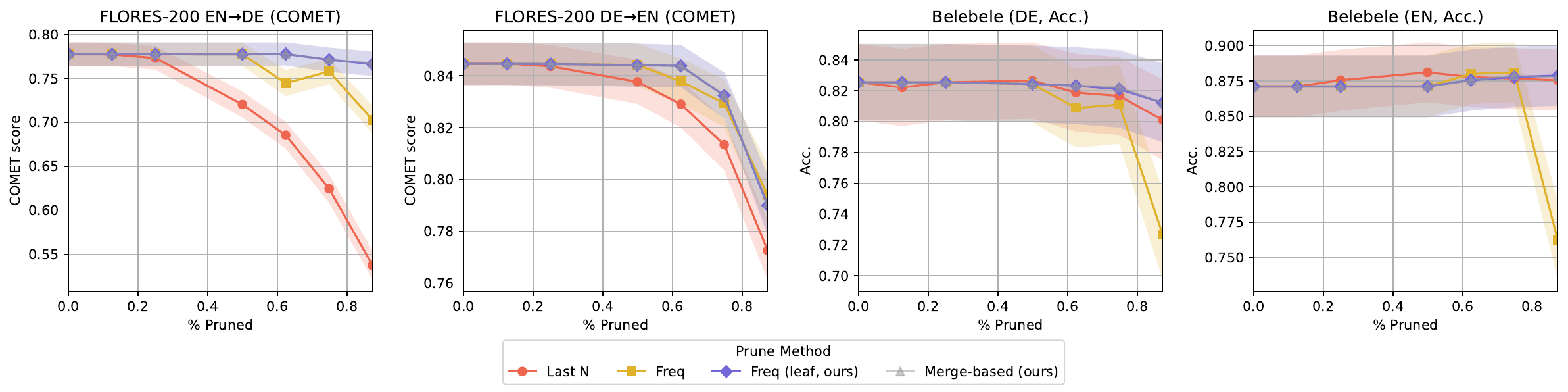}
    \caption{Llama-3.1-8B vocabulary pruned with different methods and evaluated on German and English downstream tasks. The dataset for training the pruner consisted of 50-50 German-English mix. The highlighted area shows the 95\% confidence interval.  Note that \textit{Freq (leaf)} and \textit{Merge-based} method results overlap.}
    \label{fig:pruning_downstream_de}
\end{figure*}

\begin{figure*}[h!]
    \centering \includegraphics[width=0.90\textwidth]{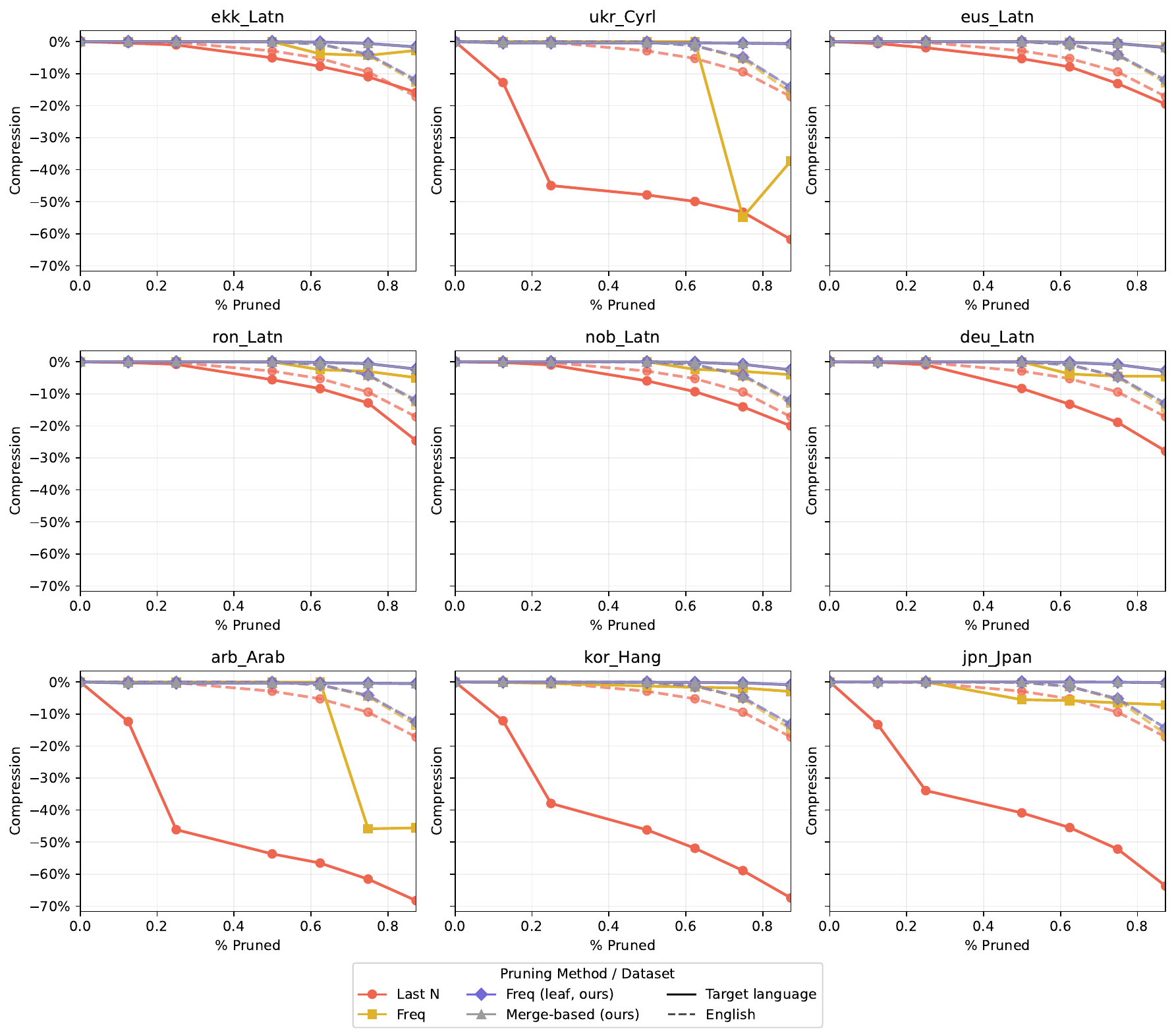}
    \caption{Change in \textbf{compression} (\textit{bytes per token} $\uparrow$) depending on the pruning method. Note that \textit{Freq (leaf)} and \textit{Merge-based} method results overlap.}
\label{fig:pruning_figure_multi_comp}
\end{figure*}

\begin{figure*}[h!]
    \centering \includegraphics[width=0.90\textwidth]{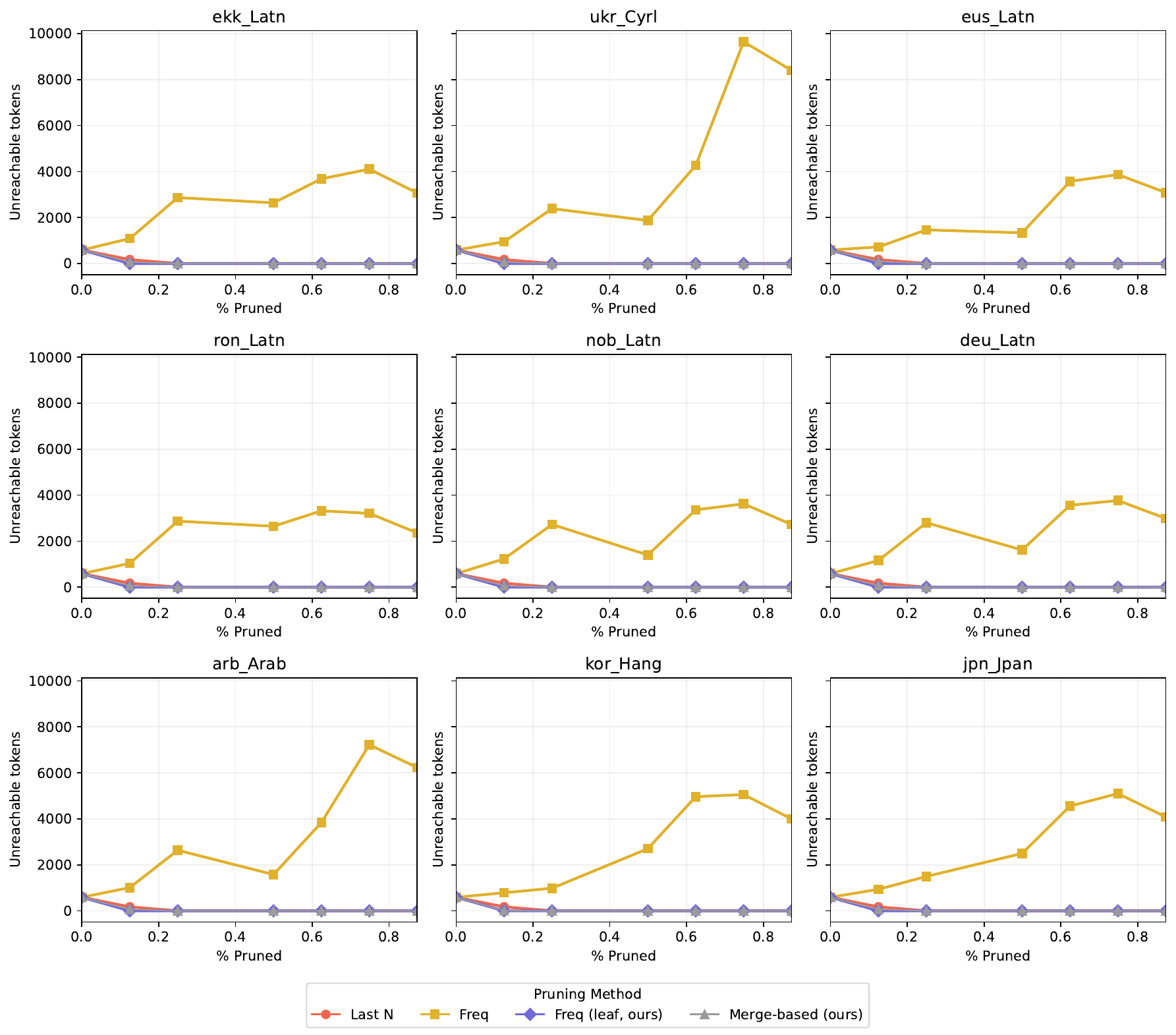}
    \caption{Number of \textbf{unreachable tokens$\downarrow$} created by different pruning methods.}
\label{fig:pruning_figure_multi_unreach}
\end{figure*}

We provide pruning compression and unreachable token curves for additional languages in Figures~\ref{fig:pruning_figure_multi_comp}~and~\ref{fig:pruning_figure_multi_unreach}, respectively.
Additionally, we provide downstream evaluation of pruning for German in Figure~\ref{fig:pruning_downstream_de}, where our pruning methods perform the best.

\newpage
\subsection{Full tokenizer extension results}
\label{sec:full-70-lang}
We provide the full improvement of our method (continued BPE training) over the naive method in Tables~\ref{tab:compression-gain-full-70}~and~~\ref{tab:compression-gain-full-70-2} for compression and in Table~\ref{tab:unreach-reduction-full-70}~and~\ref{tab:unreach-reduction-full-70-2} for unreachable tokens. We list full compression values for continued BPE training in Tables~\ref{tab:full-bpt-70}~and~\ref{tab:full-bpt-70-2}.

\begin{table*}[!hp]\centering
\scriptsize
\begin{tabular}{lcccccccccccccc}\toprule
&\multicolumn{6}{c}{\textbf{Llama-3 extension}} & &\multicolumn{6}{c}{\textbf{Qwen-2.5 extension}} \\\cmidrule{2-7}\cmidrule{9-14}
\textbf{Lang.} &\textbf{1000} &\textbf{2000} &\textbf{4000} &\textbf{8000} &\textbf{16000} &\textbf{32000} & &\textbf{1000} &\textbf{2000} &\textbf{4000} &\textbf{8000} &\textbf{16000} &\textbf{32000} \\\midrule
afr\_Latn &\cellcolor[HTML]{f9f8fe}1.9 &\cellcolor[HTML]{f8f8fd}2.1 &\cellcolor[HTML]{f7f7fd}2.3 &\cellcolor[HTML]{f9f9fe}1.7 &\cellcolor[HTML]{fbfafe}1.4 &\cellcolor[HTML]{fbfbfe}1.2 & &\cellcolor[HTML]{f3f2fc}3.4 &\cellcolor[HTML]{f0effb}4.4 &\cellcolor[HTML]{ebeafa}5.6 &\cellcolor[HTML]{e8e6f9}6.6 &\cellcolor[HTML]{e6e4f8}7.2 &\cellcolor[HTML]{e6e4f8}7.2 \\
als\_Latn &\cellcolor[HTML]{eeedfb}5.0 &\cellcolor[HTML]{efeefb}4.6 &\cellcolor[HTML]{f1f0fb}4.1 &\cellcolor[HTML]{f2f1fc}3.9 &\cellcolor[HTML]{f6f6fd}2.6 &\cellcolor[HTML]{f9f8fe}1.9 & &\cellcolor[HTML]{e3e1f8}8.0 &\cellcolor[HTML]{dedbf6}9.5 &\cellcolor[HTML]{d9d6f5}10.8 &\cellcolor[HTML]{d9d6f5}11.0 &\cellcolor[HTML]{dcdaf6}10.0 &\cellcolor[HTML]{dcd9f6}10.1 \\
arb\_Arab &\cellcolor[HTML]{fcfcff}0.9 &\cellcolor[HTML]{fcfcff}1.0 &\cellcolor[HTML]{fcfbfe}1.1 &\cellcolor[HTML]{fbfbfe}1.3 &\cellcolor[HTML]{fcfcff}0.9 &\cellcolor[HTML]{fdfdff}0.6 & &\cellcolor[HTML]{faf9fe}1.5 &\cellcolor[HTML]{f8f8fd}2.0 &\cellcolor[HTML]{f5f4fd}2.8 &\cellcolor[HTML]{f4f3fc}3.2 &\cellcolor[HTML]{f3f2fc}3.5 &\cellcolor[HTML]{f2f1fc}3.8 \\
ary\_Arab &\cellcolor[HTML]{fefeff}0.4 &\cellcolor[HTML]{fcfcff}0.9 &\cellcolor[HTML]{fcfcff}0.9 &\cellcolor[HTML]{fcfcff}0.9 &\cellcolor[HTML]{fcfcff}0.9 &\cellcolor[HTML]{fdfdff}0.6 & &\cellcolor[HTML]{fdfdff}0.6 &\cellcolor[HTML]{fafafe}1.3 &\cellcolor[HTML]{fafafe}1.5 &\cellcolor[HTML]{f9f9fe}1.7 &\cellcolor[HTML]{f7f6fd}2.3 &\cellcolor[HTML]{f7f6fd}2.4 \\
\bottomrule
\end{tabular}
\caption{Gain in \textbf{compression} (\%) of \textbf{continued tokenizer training} over\textbf{ naive extension} from an independent tokenizer. Continues in Table~\ref{tab:compression-gain-full-70-2}.}\label{tab:compression-gain-full-70}
\end{table*}

\begin{table*}[!htp]\centering
\scriptsize
\begin{tabular}{lcccccccccccccc}\toprule
&\multicolumn{6}{c}{\textbf{Mistral Nemo}} & &\multicolumn{6}{c}{\textbf{Llama-2}} \\\cmidrule{2-7}\cmidrule{9-14}
\textbf{Lang.} &\textbf{1000} &\textbf{2000} &\textbf{4000} &\textbf{8000} &\textbf{16000} &\textbf{32000} & &\textbf{1000} &\textbf{2000} &\textbf{4000} &\textbf{8000} &\textbf{16000} &\textbf{32000} \\\midrule
afr\_Latn &\cellcolor[HTML]{fafafe}1.6 &\cellcolor[HTML]{f9f9fe}1.8 &\cellcolor[HTML]{faf9fe}1.7 &\cellcolor[HTML]{fbfafe}1.4 &\cellcolor[HTML]{fbfbfe}1.2 &\cellcolor[HTML]{fcfcff}1.0 & &\cellcolor[HTML]{f5f4fd}3.0 &\cellcolor[HTML]{f1f0fb}4.1 &\cellcolor[HTML]{eeedfb}5.0 &\cellcolor[HTML]{eae9fa}6.0 &\cellcolor[HTML]{e8e6f9}6.6 &\cellcolor[HTML]{e8e6f9}6.7 \\
als\_Latn &\cellcolor[HTML]{f4f3fc}3.3 &\cellcolor[HTML]{f5f4fd}2.9 &\cellcolor[HTML]{f5f4fd}2.9 &\cellcolor[HTML]{f5f4fd}3.0 &\cellcolor[HTML]{f7f7fd}2.3 &\cellcolor[HTML]{faf9fe}1.6 & &\cellcolor[HTML]{e3e1f8}8.0 &\cellcolor[HTML]{dfdcf6}9.3 &\cellcolor[HTML]{dbd9f5}10.3 &\cellcolor[HTML]{d7d4f4}11.5 &\cellcolor[HTML]{dbd8f5}10.4 &\cellcolor[HTML]{dbd9f5}10.3 \\
arb\_Arab &0.2 &\cellcolor[HTML]{fffeff}0.3 &0.2 &\cellcolor[HTML]{fffeff}0.3 &\cellcolor[HTML]{fefeff}0.3 &\cellcolor[HTML]{fefeff}0.3 & &\cellcolor[HTML]{fafafe}1.6 &\cellcolor[HTML]{fdfdff}0.7 &\cellcolor[HTML]{fefeff}0.3 &0.1 &\cellcolor[HTML]{fefefe}-0.2 &\cellcolor[HTML]{fefefe}-0.2 \\
ary\_Arab &\cellcolor[HTML]{fefeff}0.4 &\cellcolor[HTML]{fefefe}-0.0 &0.0 &0.1 &0.1 &\cellcolor[HTML]{fefeff}0.3 & &\cellcolor[HTML]{fdfcff}0.8 &\cellcolor[HTML]{fefeff}0.4 &0.2 &0.1 &\cellcolor[HTML]{fefefe}-0.1 &\cellcolor[HTML]{fefefe}-0.2 \\
azj\_Latn &\cellcolor[HTML]{f3f2fc}3.6 &\cellcolor[HTML]{f0eefb}4.5 &\cellcolor[HTML]{edecfa}5.2 &\cellcolor[HTML]{ecebfa}5.5 &\cellcolor[HTML]{edecfa}5.1 &\cellcolor[HTML]{f0effb}4.3 & &\cellcolor[HTML]{d6d3f4}11.8 &\cellcolor[HTML]{cfcbf2}13.8 &\cellcolor[HTML]{c0bcee}18.0 &\cellcolor[HTML]{b8b3ec}20.4 &\cellcolor[HTML]{b1abea}22.4 &\cellcolor[HTML]{b2acea}22.1 \\
\bottomrule
\end{tabular}
\caption{Gain in \textbf{compression} (\%) of \textbf{continued tokenizer training} over\textbf{ naive extension} from an independent tokenizer.}\label{tab:compression-gain-full-70-2}
\end{table*}

\begin{table*}[!htp]\centering
\scriptsize
\begin{tabular}{lcccccccccccccc}\toprule
&\multicolumn{6}{c}{\textbf{Llama-3 extension}} & &\multicolumn{6}{c}{\textbf{Qwen-2.5 extension}} \\\cmidrule{2-7}\cmidrule{9-14}
\textbf{Lang.} &\textbf{1000} &\textbf{2000} &\textbf{4000} &\textbf{8000} &\textbf{16000} &\textbf{32000} & &\textbf{1000} &\textbf{2000} &\textbf{4000} &\textbf{8000} &\textbf{16000} &\textbf{32000} \\\midrule
afr\_Latn &\cellcolor[HTML]{fefefe}-74 &\cellcolor[HTML]{fdfdfe}-156 &\cellcolor[HTML]{fafafd}-357 &\cellcolor[HTML]{f4f4fc}-875 &\cellcolor[HTML]{e9e7f9}-1858 &\cellcolor[HTML]{d4d1f3}-3692 & &\cellcolor[HTML]{fefefe}-72 &\cellcolor[HTML]{fdfdfe}-155 &\cellcolor[HTML]{fafafd}-350 &\cellcolor[HTML]{f4f4fc}-882 &\cellcolor[HTML]{e9e7f9}-1859 &\cellcolor[HTML]{d4d1f3}-3676 \\
als\_Latn &\cellcolor[HTML]{fdfdfe}-95 &\cellcolor[HTML]{fcfcfe}-229 &\cellcolor[HTML]{f9f9fd}-478 &\cellcolor[HTML]{f3f2fb}-983 &\cellcolor[HTML]{e7e6f8}-2004 &\cellcolor[HTML]{cfccf1}-4071 & &\cellcolor[HTML]{fdfdfe}-94 &\cellcolor[HTML]{fcfcfe}-231 &\cellcolor[HTML]{f9f8fd}-491 &\cellcolor[HTML]{f3f2fb}-1007 &\cellcolor[HTML]{e7e5f8}-2045 &\cellcolor[HTML]{cfcbf1}-4127 \\
arb\_Arab &\cellcolor[HTML]{fefefe}-23 &\cellcolor[HTML]{fefefe}-58 &\cellcolor[HTML]{fdfdfe}-141 &\cellcolor[HTML]{fbfbfe}-293 &\cellcolor[HTML]{f8f8fd}-547 &\cellcolor[HTML]{f2f2fb}-1039 & &\cellcolor[HTML]{fefefe}-15 &\cellcolor[HTML]{fefefe}-52 &\cellcolor[HTML]{fdfcfe}-162 &\cellcolor[HTML]{fafafd}-366 &\cellcolor[HTML]{f6f5fc}-733 &\cellcolor[HTML]{eeedfa}-1442 \\
ary\_Arab &\cellcolor[HTML]{fefefe}-27 &\cellcolor[HTML]{fefefe}-52 &\cellcolor[HTML]{fdfdfe}-120 &\cellcolor[HTML]{fbfbfe}-264 &\cellcolor[HTML]{f9f8fd}-510 &\cellcolor[HTML]{f3f3fb}-954 & &\cellcolor[HTML]{fefefe}-28 &\cellcolor[HTML]{fefefe}-63 &\cellcolor[HTML]{fdfcfe}-163 &\cellcolor[HTML]{fafafd}-358 &\cellcolor[HTML]{f6f5fc}-728 &\cellcolor[HTML]{eeedfa}-1411 \\
\bottomrule
\end{tabular}
\caption{\textbf{Change in unreachable tokens through merges} of \textbf{continued tokenizer training} compared to \textbf{naive extension} from an independent tokenizer. Continues in Table~\ref{tab:unreach-reduction-full-70-2}.}\label{tab:unreach-reduction-full-70}
\end{table*}

\begin{table*}[!htp]\centering
\scriptsize
\begin{tabular}{lcccccccccccccc}\toprule
&\multicolumn{6}{c}{\textbf{Mistral Nemo extension}} & &\multicolumn{6}{c}{\textbf{Llama-2 extension}} \\\cmidrule{2-7}\cmidrule{9-14}
\textbf{Lang.} &\textbf{1000} &\textbf{2000} &\textbf{4000} &\textbf{8000} &\textbf{16000} &\textbf{32000} & &\textbf{1000} &\textbf{2000} &\textbf{4000} &\textbf{8000} &\textbf{16000} &\textbf{32000} \\\midrule
afr\_Latn &\cellcolor[HTML]{fefefe}-62 &\cellcolor[HTML]{fdfdfe}-132 &\cellcolor[HTML]{fbfafd}-329 &\cellcolor[HTML]{f5f5fc}-793 &\cellcolor[HTML]{eceaf9}-1634 &\cellcolor[HTML]{d8d6f4}-3290 & &\cellcolor[HTML]{fefefe}-59 &\cellcolor[HTML]{fdfdfe}-149 &\cellcolor[HTML]{fbfafd}-330 &\cellcolor[HTML]{f6f5fc}-753 &\cellcolor[HTML]{eceaf9}-1617 &\cellcolor[HTML]{dad7f4}-3187 \\
als\_Latn &\cellcolor[HTML]{fefdfe}-83 &\cellcolor[HTML]{fcfcfe}-197 &\cellcolor[HTML]{f9f9fd}-437 &\cellcolor[HTML]{f4f4fc}-880 &\cellcolor[HTML]{e9e8f9}-1833 &\cellcolor[HTML]{d4d1f3}-3661 & &\cellcolor[HTML]{fefefe}-78 &\cellcolor[HTML]{fcfcfe}-226 &\cellcolor[HTML]{f9f8fd}-514 &\cellcolor[HTML]{f2f1fb}-1070 &\cellcolor[HTML]{e6e5f8}-2074 &\cellcolor[HTML]{cecaf1}-4218 \\
arb\_Arab &\cellcolor[HTML]{fefefe}-9 &\cellcolor[HTML]{fefefe}-28 &\cellcolor[HTML]{fefefe}-58 &\cellcolor[HTML]{fdfdfe}-114 &\cellcolor[HTML]{fcfbfe}-245 &\cellcolor[HTML]{f8f7fd}-577 & &0 &0 &0 &\cellcolor[HTML]{fefefe}-2 &\cellcolor[HTML]{fefefe}-4 &\cellcolor[HTML]{fefefe}-12 \\
ary\_Arab &\cellcolor[HTML]{fefefe}-10 &\cellcolor[HTML]{fefefe}-24 &\cellcolor[HTML]{fefefe}-55 &\cellcolor[HTML]{fdfdfe}-103 &\cellcolor[HTML]{fbfbfe}-264 &\cellcolor[HTML]{f8f8fd}-550 & &0 &0 &0 &\cellcolor[HTML]{fefefe}-1 &\cellcolor[HTML]{fefefe}-1 &\cellcolor[HTML]{fefefe}-6 \\
\bottomrule
\end{tabular}
\caption{\textbf{Change in unreachable tokens through merges} of \textbf{continued tokenizer training} compared to \textbf{naive extension} from an independent tokenizer.}\label{tab:unreach-reduction-full-70-2}
\end{table*}

\begin{table*}[!htp]\centering
\addtolength{\tabcolsep}{-0.7pt}
\scriptsize
\begin{tabular}{lcccccccccccccccc}\toprule
&\multicolumn{7}{c}{\textbf{Llama-3 extension}} & &\multicolumn{7}{c}{\textbf{Qwen-2.5 extension}} \\\cmidrule{2-8}\cmidrule{10-16}
\textbf{Lang.} &\textbf{0} &\textbf{1000} &\textbf{2000} &\textbf{4000} &\textbf{8000} &\textbf{16000} &\textbf{32000} & &\textbf{0} &\textbf{1000} &\textbf{2000} &\textbf{4000} &\textbf{8000} &\textbf{16000} &\textbf{32000} \\\midrule
afr\_Latn &\cellcolor[HTML]{eceafa}3.093 &\cellcolor[HTML]{e6e5f9}3.630 &\cellcolor[HTML]{e5e3f8}3.828 &\cellcolor[HTML]{e2e0f7}4.062 &\cellcolor[HTML]{e0def7}4.301 &\cellcolor[HTML]{dedbf6}4.529 &\cellcolor[HTML]{dcdaf6}4.720 & &\cellcolor[HTML]{ecebfa}3.051 &\cellcolor[HTML]{e7e5f9}3.579 &\cellcolor[HTML]{e5e3f8}3.773 &\cellcolor[HTML]{e3e1f8}4.004 &\cellcolor[HTML]{e1def7}4.239 &\cellcolor[HTML]{dedcf6}4.463 &\cellcolor[HTML]{dddaf6}4.649 \\
als\_Latn &\cellcolor[HTML]{f0effb}2.625 &\cellcolor[HTML]{e7e5f9}3.550 &\cellcolor[HTML]{e5e3f8}3.770 &\cellcolor[HTML]{e3e0f7}4.036 &\cellcolor[HTML]{e0def7}4.328 &\cellcolor[HTML]{dddaf6}4.645 &\cellcolor[HTML]{dad7f5}4.936 & &\cellcolor[HTML]{f0effb}2.579 &\cellcolor[HTML]{e8e6f9}3.484 &\cellcolor[HTML]{e6e4f8}3.710 &\cellcolor[HTML]{e3e1f8}3.976 &\cellcolor[HTML]{e0def7}4.266 &\cellcolor[HTML]{dddbf6}4.577 &\cellcolor[HTML]{dbd8f5}4.866 \\
arb\_Arab &\cellcolor[HTML]{dcdaf6}4.705 &\cellcolor[HTML]{d8d5f5}5.138 &\cellcolor[HTML]{d5d2f4}5.438 &\cellcolor[HTML]{d1cef3}5.867 &\cellcolor[HTML]{ccc9f1}6.385 &\cellcolor[HTML]{c7c3f0}6.941 &\cellcolor[HTML]{c1bdee}7.542 & &\cellcolor[HTML]{dcdaf6}4.697 &\cellcolor[HTML]{d8d5f5}5.158 &\cellcolor[HTML]{d5d2f4}5.431 &\cellcolor[HTML]{d1cef3}5.838 &\cellcolor[HTML]{cdc9f1}6.313 &\cellcolor[HTML]{c8c4f0}6.855 &\cellcolor[HTML]{c2beef}7.438 \\
\bottomrule
\end{tabular}
\caption{Compression (\textit{bytes per token}) on FLORES for tokenizer extension with continued BPE training. The columns represent different number of added tokens where 0 is the original tokenizer. Continues is Table~\ref{tab:full-bpt-70-2}.}\label{tab:full-bpt-70}
\end{table*}

\begin{table*}[!htp]\centering
\addtolength{\tabcolsep}{-0.7pt}
\scriptsize
\begin{tabular}{lcccccccccccccccc}\toprule
&\multicolumn{7}{c}{\textbf{Mistral Nemo}} & &\multicolumn{7}{c}{\textbf{Llama-2}} \\\cmidrule{2-8}\cmidrule{10-16}
\textbf{Lang.} &\textbf{0} &\textbf{1000} &\textbf{2000} &\textbf{4000} &\textbf{8000} &\textbf{16000} &\textbf{32000} & &\textbf{0} &\textbf{1000} &\textbf{2000} &\textbf{4000} &\textbf{8000} &\textbf{16000} &\textbf{32000} \\\midrule
afr\_Latn &\cellcolor[HTML]{e9e8f9}3.343 &\cellcolor[HTML]{e5e4f8}3.728 &\cellcolor[HTML]{e4e2f8}3.886 &\cellcolor[HTML]{e2e0f7}4.087 &\cellcolor[HTML]{e0def7}4.294 &\cellcolor[HTML]{dedcf6}4.499 &\cellcolor[HTML]{dddaf6}4.666 & &\cellcolor[HTML]{edecfa}2.932 &\cellcolor[HTML]{e8e6f9}3.461 &\cellcolor[HTML]{e6e4f8}3.660 &\cellcolor[HTML]{e4e2f8}3.903 &\cellcolor[HTML]{e2dff7}4.145 &\cellcolor[HTML]{dfddf7}4.379 &\cellcolor[HTML]{dddbf6}4.580 \\
als\_Latn &\cellcolor[HTML]{ecebfa}3.049 &\cellcolor[HTML]{e7e5f9}3.613 &\cellcolor[HTML]{e5e3f8}3.817 &\cellcolor[HTML]{e2e0f7}4.061 &\cellcolor[HTML]{e0ddf7}4.332 &\cellcolor[HTML]{dddaf6}4.623 &\cellcolor[HTML]{dad8f5}4.887 & &\cellcolor[HTML]{f3f3fc}2.259 &\cellcolor[HTML]{e9e7f9}3.384 &\cellcolor[HTML]{e7e5f9}3.616 &\cellcolor[HTML]{e4e2f8}3.887 &\cellcolor[HTML]{e1dff7}4.188 &\cellcolor[HTML]{dedcf6}4.503 &\cellcolor[HTML]{dbd9f5}4.797 \\
arb\_Arab &\cellcolor[HTML]{cfcbf2}6.115 &\cellcolor[HTML]{cecaf2}6.232 &\cellcolor[HTML]{cdc9f1}6.333 &\cellcolor[HTML]{cbc7f1}6.501 &\cellcolor[HTML]{c8c4f0}6.786 &\cellcolor[HTML]{c5c0ef}7.182 &\cellcolor[HTML]{c0bcee}7.643 & &\cellcolor[HTML]{f6f5fd}1.988 &\cellcolor[HTML]{e1def7}4.248 &\cellcolor[HTML]{dbd9f6}4.780 &\cellcolor[HTML]{d5d2f4}5.432 &\cellcolor[HTML]{cecbf2}6.144 &\cellcolor[HTML]{c8c4f0}6.845 &\cellcolor[HTML]{c1bdee}7.529 \\
\bottomrule
\end{tabular}
\caption{Compression (\textit{bytes per token}) on FLORES for tokenizer extension with continued BPE training. The columns represent different number of added tokens where 0 is the original tokenizer.}\label{tab:full-bpt-70-2}
\end{table*}

\end{document}